\documentclass[final]{cvpr}

\usepackage{xspace}
\usepackage{times}
\usepackage{epsfig}
\usepackage{graphicx}
\usepackage{amsmath}
\usepackage{amssymb}

\usepackage{bm} 
\usepackage{bbm} 
\usepackage{booktabs} 
\usepackage{multirow} 
\usepackage[ruled,vlined,linesnumbered]{algorithm2e} 
\usepackage{textcomp} 
\usepackage{siunitx} 
\usepackage{paralist} 

\newcommand{\argmin}{\mathop{\mathrm{argmin}}} 
\newcommand{\argmax}{\mathop{\mathrm{argmax}}} 

\usepackage{subcaption} 
\usepackage{xspace}
\usepackage{array} 
\newcommand{\modelname}{ProtoTree\xspace}
\newcommand{\modelnames}{ProtoTrees\xspace}
\newcommand{\fullmodelname}{Neural Prototype Tree\xspace}

\usepackage{stfloats} 
\usepackage[breaklinks=true,bookmarks=false]{hyperref}

\def\cvprPaperID{05419} 
\def\confYear{CVPR 2021}
\begin{document}

\title{Neural Prototype Trees for Interpretable Fine-grained Image Recognition}

\author{Meike Nauta\textsuperscript{1} \quad Ron van Bree\textsuperscript{1} \quad Christin Seifert\textsuperscript{1,2}\\
\textsuperscript{1} University of Twente, the Netherlands \quad
\textsuperscript{2} University of Duisburg-Essen, Germany\\
{\tt\small m.nauta@utwente.nl}, {\tt\small r.j.vanbree@student.utwente.nl}, {\tt\small christin.seifert@uni-due.de}
}

\twocolumn[{%
\renewcommand\twocolumn[1][]{#1}%
\maketitle
\thispagestyle{empty}
\begin{center}
    \centering
    \includegraphics[width=0.95\textwidth,height=4.64cm]{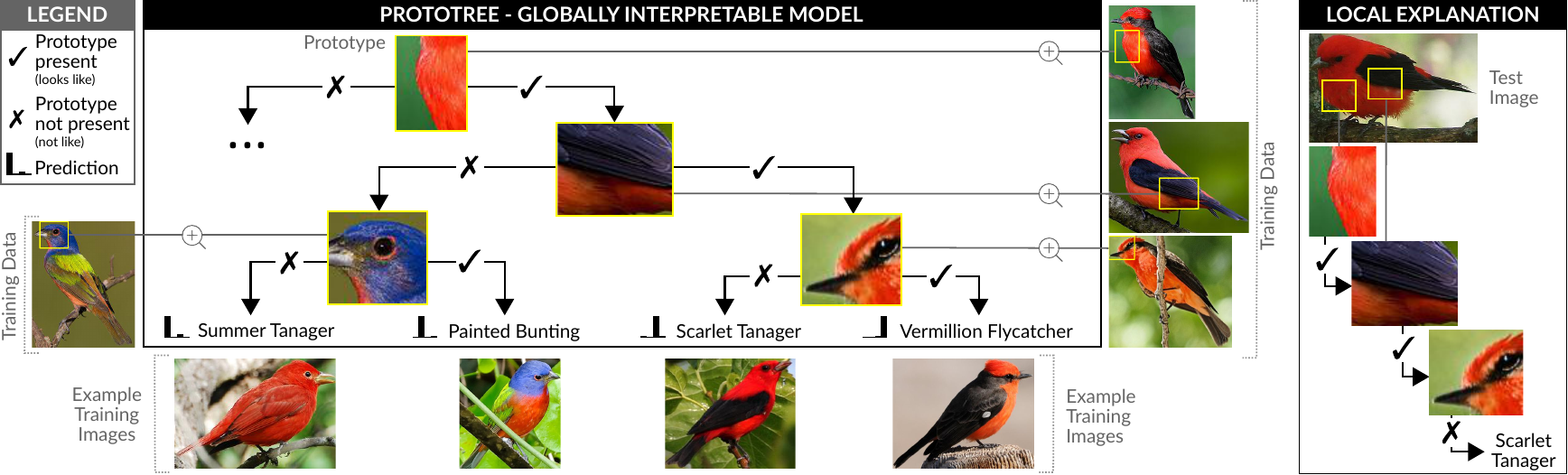}
    \captionof{figure}{A \modelname is a globally interpretable model faithfully explaining its entire reasoning (left, partially shown). Additionally, the decision making process for a single prediction can be followed (right): the presence of a red chest and black wing, and the absence of a black stripe near the eye, identifies a Scarlet Tanager. A pruned \modelname learns roughly 200 prototypes for CUB (dataset with 200 bird species), making only 8 local decisions on average for one test image.
    }
    \label{fig:prototree_example_global_local}
\end{center}%
}]

\begin{abstract}
Prototype-based methods use interpretable representations to address the black-box nature of deep learning models, in contrast to post-hoc explanation methods that only approximate such models.
We propose the Neural Prototype Tree (ProtoTree), an intrinsically interpretable deep learning method for fine-grained image recognition. ProtoTree combines prototype learning with decision trees, and thus results in a globally interpretable model by design. 
Additionally, ProtoTree can locally explain a single prediction by outlining a decision path through the tree. Each node in our binary tree contains a trainable prototypical part. The presence or absence of this learned prototype in an image determines the routing through a node. Decision making is therefore similar to human reasoning: Does the bird have a red throat? And an elongated beak? Then it’s a hummingbird! We tune the accuracy-interpretability trade-off using ensemble methods, pruning and binarizing. We apply pruning without sacrificing accuracy, resulting in a small tree with only 8 learned prototypes along a path to classify a bird from 200 species. An ensemble of 5 ProtoTrees achieves competitive accuracy on the CUB-200- 2011 and Stanford Cars data sets. Code is available at \url{github.com/M-Nauta/ProtoTree}.
\end{abstract}

\section{Introduction}
There is an ongoing scientific dispute between simple, interpretable models and complex black boxes, such as Deep Neural Networks (DNNs). DNNs have achieved superior performance, especially in computer vision, 
but their complex architectures and high-dimensional feature spaces has led to an increasing demand for transparency, interpretability and explainability~\cite{adadi2018peeking}, particularly in domains with high-stakes decisions~\cite{rudin2019stop}.
In contrast, decision trees are easy to understand and interpret~\cite{comprehensible_freitas,guidotti2018survey}, because they
transparently arrange decision rules in a hierarchical structure. 
Their predictive performance is however far from competitive for 
computer vision tasks. We address this so-called `accuracy-interpretability trade-off'~\cite{adadi2018peeking,Lipton2018mythos} by combining the expressiveness of deep learning with the interpretability of decision trees.

We present the \emph{\fullmodelname}, \modelname in short, an intrinsically interpretable method for fine-grained image recognition. A \modelname has the representational power of a neural network, and contains a built-in binary decision tree structure, as shown in Fig.~\ref{fig:prototree_example_global_local} (left). 
Each internal node in the tree contains a trainable \emph{prototype}. 
Our prototypes are prototypical \emph{parts} learned with backpropagation, as introduced in the Prototypical Part Network (ProtoPNet)~\cite{NIPS2019_protopnet} where a prototype is a trainable tensor that can be visualized as a patch of a training sample. 
The extent to which this prototype is present in an input image determines the routing of the image through the corresponding node. Leaves of the \modelname learn class distributions. The paths from root to leaves represent the learned classification rules.
The reasoning of our model is thus similar to the ``Guess Who?'' game where a player asks a series of binary questions related to visual properties to find out which of the 24 displayed images the other player had in mind.

To this end, a \modelname consists of a Convolutional Neural Network (CNN) followed by a binary tree structure and can be trained end-to-end with a standard cross-entropy loss function. We only require class labels and do not need any other annotations.
To make the tree differentiable and back-propagation compatible, we utilize a \emph{soft} decision tree, meaning that a sample is routed through both children, each with a certain weight. We present a novel routing procedure based on the similarity between the latent image embedding and a prototype. 
We show that a trained soft \modelname can be converted to a hard, and therefore more interpretable, \modelname without loss of accuracy. 

A \modelname approximates the accuracy of non-interpretable classifiers, while being \emph{interpretable-by-design} and offering truthful global and local explanations. This way it provides a novel take on interpretable machine learning. 
In contrast to \emph{post-hoc} explanations, which approximate a trained model or its output~\cite{montavon2018methods,laugel2019dangers}, a \modelname is inherently interpretable since it directly incorporates interpretability in the structure of the predictive model~\cite{montavon2018methods}.  
A \modelname therefore faithfully shows its entire classification behaviour, independent of its input, providing a \emph{global} explanation (Fig.~\ref{fig:prototree_example_global_local}). As a consequence, our compact tree enables a human to convey, or even print out, the \emph{whole} model.
In contrast to \emph{local} explanations, which explain a single prediction and can be unstable and contradicting~\cite{alvarez2018robustness,Kindermans2019}, global explanations enable \emph{simulatability}~\cite{Lipton2018mythos}.
Additionally, our \modelname can produce \emph{local} explanations by showing the routing of a specific input image through the tree (Fig.~\ref{fig:prototree_example_global_local}, right). Hence, ProtoTree allows retraceable decisions in a human-comprehensible number of steps. In case of a misclassification, the responsible node can be easily identified by tracking down the series of decisions, which eases error analysis.

\noindent\textbf{Scientific Contributions}
\begin{compactitem}
\item An intrinsically interpretable neural prototype tree architecture for fine-grained image recognition.
\item Outperforming ProtoPNet~\cite{NIPS2019_protopnet} while having roughly only 10\% of the number of prototypes, included in a built-in hierarchical structure.
\item An ensemble of 5 interpretable \modelnames achieves competitive performance on CUB-200-2011~\cite{WahCUB_200_2011} (CUB) and Stanford Cars~\cite{cars_dataset_paper}.
\end{compactitem}

\section{Related Work}
Within computer vision, various explainability strategies exist for different notions of interpretability. A machine learning model can be explained for a single prediction, \eg part-based methods~\cite{Zheng_2017_ICCV,Zhou_2018_ECCV},
saliency maps~\cite{bach_pixelwise,Fong_2017_ICCV,Zhou_2016_CVPR} or representer points~\cite{NIPS2018_representerpoints}. Others explain the internals of a model, with \eg activation maximization~\cite{NIPS2016_6519,olah2017feature} to visualize neurons, deconvolution or upconvolution~\cite{Dosovitskiy_2016_CVPR, zeiler_deconv} to explain layers, generating image exemplars~\cite{abele_exemplars} to explain the latent space, or concept activation vectors~\cite{pmlr-v80-kim18d} to explain model sensitivity.
While such post-hoc methods give an intuition about the black-box model, intrinsic interpretable models such as classical decision trees, are fully simulatable since they faithfully show the decision making \emph{process}. Similarly, by utilizing interpretable features as splitting criteria, 
a ProtoTree’s decision making process can be understood in its entirety, as well as for a single prediction.
ProtoTree combines prototypical feature representations (Sec.~\ref{sec:related_work:prototypes}) with soft-decision tree learning (Sec.~\ref{sec:relatedwork_softdt}).

\subsection{Interpretability with Prototypes}
\label{sec:related_work:prototypes}
\begin{figure}
    \centering
    \includegraphics[width=\linewidth]{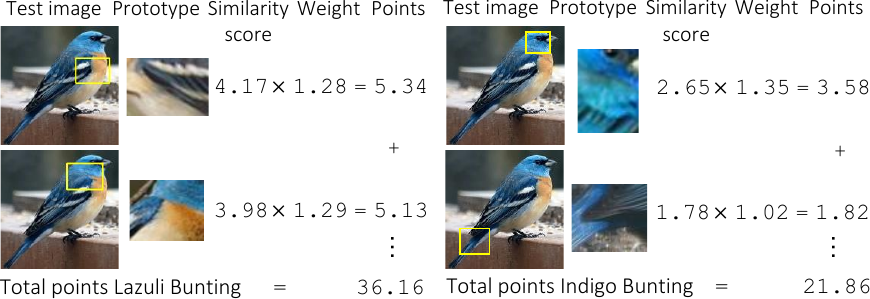}
    \caption{Excerpt from classification process of ProtoPNet~\cite{NIPS2019_protopnet}. ProtoPNet learns 10 prototypes per class, resulting in 2000 prototypes for CUB, therefore making 2000 local decisions for one test image.}
    \label{fig:protopnet_reasoning}
\end{figure}
Prototypes are visual explanations that can be incorporated in a model for intrinsic interpretability. ProtoAttend~\cite{arik2019protoattend} and DMR~\cite{angelov2020towards} use full images. In contrast, we go for prototypical \emph{parts} to break up the decision making process in small steps. Our choice is supported by BagNet~\cite{bagnet_iclr} which found that ``even complex perceptual tasks like ImageNet can be solved just based on small image features and without any notion of spatial relationships". 
Related is the Classification-By-Components network~\cite{saralajew_NIPS2019} that learns positive, negative and indefinite visual components motivated by the recognition-by-components theory~\cite{biederman1987recognition} describing how humans recognize objects by segmenting it into multiple components.  

We build upon the Prototypical Part Network (ProtoPNet)~\cite{NIPS2019_protopnet}, an intrinsically interpretable deep network architecture for case-based reasoning. Since their prototypes have smaller spatial dimensions than the image itself, they represent prototypical \emph{parts} and are therefore suited for fine-grained image classification. ProtoPNet learns a pre-determined number of prototypical parts (prototypes) \emph{per class}. To classify an image, the similarity between a prototype and a patch in the image is calculated by measuring the distance in latent space.
The resulting similarity scores are weighted by values learned by a fully-connected layer. 
The explanation of ProtoPNet shows the reasoning process for a single image, by visualizing all prototypes together with their weighted similarity score. Summing the weighted similarity scores per class gives a final score for the image belonging to each class, as shown in Fig.~\ref{fig:protopnet_reasoning}. 
We improve upon ProtoPNet by showing an easy-to-interpret \emph{global} explanation by means of a decision tree. Such a hierarchical, logical model aids interpretability~\cite{comprehensible_freitas,rudin2019stop}, since a tree has various conceptual advantages compared to a linear bag-of-prototypes: a tree enforces a \emph{sequence} of steps and it supports negative associations (\ie absence of prototype), thereby reducing the number of prototypes and better mimicking human reasoning. The hierarchical structure therefore enhances interpretability and could also lead to more insights w.r.t. clusters in the data. 
Instead of multiplying similarity scores with weights, our local explanation shows the routing of a sample through the tree. Furthermore, we do not have class-specific prototypes, do not need to learn weights for similarity scores and we use a simple cross-entropy loss function. 

\subsection{Neural Soft Decision Trees}
\label{sec:relatedwork_softdt}
\begin{figure}
    \centering
    \begin{subfigure}[t]{0.115\textwidth}
        \centering
        \includegraphics[width=\textwidth]{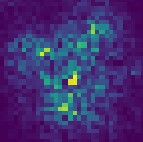}
        \caption{DNDF~\cite{li2019visualizing}}
        \label{fig:vis_dndf}
    \end{subfigure}
    \begin{subfigure}[t]{0.115\textwidth}
        \centering
        \includegraphics[width=\textwidth]{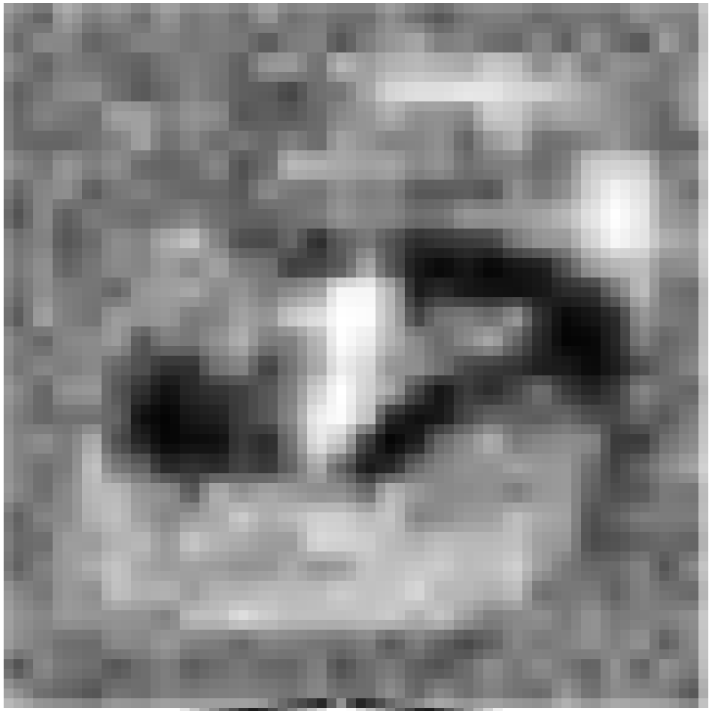}
        \caption{SDT~\cite{frosst2017distilling}}
        \label{fig:vis_sdt_frosst}
    \end{subfigure}
    \begin{subfigure}[t]{0.115\textwidth}
        \centering
        \includegraphics[width=\textwidth]{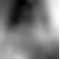}
        \caption{SDT~\cite{hehn2019end}}
        \label{fig:vis_hehn}
    \end{subfigure}
    \begin{subfigure}[t]{0.10\textwidth}
        \centering
        \includegraphics[width=\textwidth]{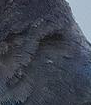}
        \caption{Ours}
        \label{fig:vis_root_own}
    \end{subfigure}
    \caption{Visualized root node from soft decision trees. Applied to resp. CIFAR10, MNIST, FashionMNIST and CUB. Republished with permission from the authors (a-c).}
    \label{fig:related_work_vis}
\end{figure}

Soft Decision Trees (SDTs) have shown to be more accurate than traditional hard decision trees~\cite{irsoy2012soft, irsoy2014budding,suarez1999globally}.
Only recently deep neural networks are integrated in binary SDTs.
The Deep Neural Decision Forest (DNDF)~\cite{Kontschieder_2015_ICCV} is an ensemble of neural SDTs: a neural network learns a latent representation of the input, and each node learns a routing function. 
Adaptive Neural Trees~\cite{ANTS_tanno19a} (ANTs) are a generalization of DNDF.
Each node can transform and route its input with a small neural network. 
In contrast to most SDTs that require a fixed tree structure, including ours, ANTs greedily learn a binary tree topology. Such greedy algorithms however could lead to suboptimal trees~\cite{norouzi2015efficient}, and are only applied to simple classification problems such as MNIST. 
Furthermore, the above methods lose the main attractive property of decision trees: interpretability.
DNDFs can be locally interpreted 
by visualizing a path of saliency maps~\cite{li2019visualizing}, as shown in Fig.~\ref{fig:vis_dndf}.
Frosst \& Hinton~\cite{frosst2017distilling} train a perceptron for each node, and visualize the learned weights (Fig.~\ref{fig:vis_sdt_frosst}). The limited representational power of perceptrons however leads to suboptimal classification results.
The approach in~\cite{hehn2019end} makes SDTs deterministic at test time and linear split parameters can be visualized and enhanced with a spatial regularization term (Fig.~\ref{fig:vis_hehn}).
In contrast to these interpretable methods, we apply our method to natural images for fine-grained image recognition and our visualizations are sharp and full-color, therefore improving interpretability (Fig.~\ref{fig:vis_root_own}). 
Instead of image recognition, Neural-Backed Decision Trees for Segmentation~\cite{wan2020segnbdt} use visual decision rules with saliency maps for segmentation.

Other tree approaches for image classification acpply post-hoc explanation techniques,
by showing example images per node \cite{alaniz2019explainable, zhang2019interpreting}, visualizing typical CNN filters of each node that can be manually labelled \cite{zhang2019interpreting}, showing class activation maps~\cite{ji2020attention} or 
manual inspection of leaf labels and the meaning of internal nodes~\cite{wan2020nbdt}. 
We extend prior work by including prototypes in a tree structure, thereby obtaining a globally explainable, \emph{intrinsically} interpretable model with only one decision per node. Additionally, similar to ProtoPNet~\cite{NIPS2019_protopnet}, a \modelname does not require manual labelling and is therefore self-explanatory.
Our work differs from hierarchical image classification (\eg, a gibbon is an animal and a primate) such as~\cite{hase2019interpretable}, since we do not require hierarchical labels or a predefined taxonomy.

\section{\fullmodelname}
A \fullmodelname (\modelname) hierarchically routes an image through a binary tree for interpretable image recognition.
We now formalise the definition of a \modelname for supervised learning. Consider a classification problem with training set $\bm{\mathcal{T}}$ containing $N$ labelled images $\{(\bm{x}^{(1)}, y^{(1)}), ..., (\bm{x}^{(N)}, y^{(N)})\} \in \mathcal{X} \times \mathcal{Y}$. Given an input image $\bm{x}$, a \modelname predicts the class probability distribution over $K$ classes, denoted as $\hat{\bm{y}}$. We use $\bm{y}$ to denote the one-hot encoded ground-truth label $y$ such that we can train a \modelname by minimizing the cross-entropy loss between $\bm{y}$ and $\hat{\bm{y}}$. A ProtoTree can also be trained with soft labels from a trained model for knowledge distillation, similar to~\cite{frosst2017distilling}. 
\begin{figure*}
\begin{center}
\includegraphics[width=\linewidth]{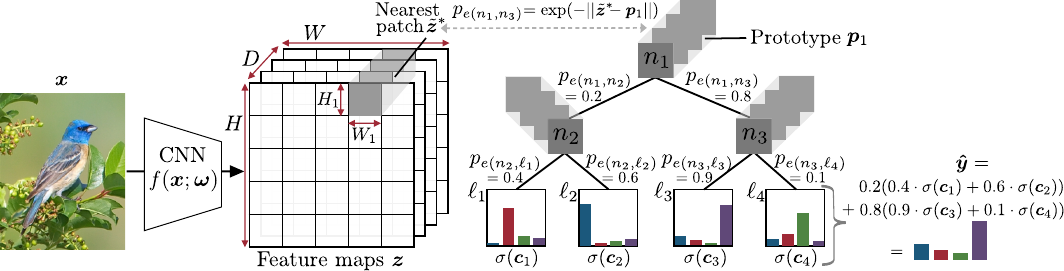}
\end{center}
   \caption{Decision making process of a \modelname to predict class probability distribution $\bm{\hat{y}}$ of input image $\bm{x}$. During training, prototypes $\bm{p}_n \in \bm{P}$, leaves' class distributions $\bm{c}$ and CNN parameters $\omega$ are learned. Probabilities $p_e$ (shown with example values) depend on the similarity between a patch in the latent input image and a prototype.}
\label{fig:prototree_reasoning_overview}
\end{figure*}

A \modelname $T$ is a combination of a convolutional neural network (CNN) $f$ with a soft neural binary decision tree structure. 
As shown in Fig.~\ref{fig:prototree_reasoning_overview}, 
an input image is first forwarded through $f$. The resulting convolutional output $\bm{z} = f(\bm{x};\bm{\omega})$ consists of $D$ two-dimensional ($H \times W$) feature maps, where $\bm{\omega}$ denotes the trainable parameters of $f$. 
Secondly, the latent representation $\bm{z} \in \mathbb{R}^{H\times W\times D}$ serves as input for a binary tree. 
This tree consists of a set of internal nodes $\mathcal{N}$, a set of leaf nodes $\mathcal{L}$, and a set of edges $\mathcal{E}$.
Each internal node $n \in \mathcal{N}$ has exactly two child nodes: $n.\mathit{left}$ connected by edge $e(n, n.\mathit{left}) \in \mathcal{E}$ and $n.\mathit{right}$ connected by $e(n, n.\mathit{right}) \in \mathcal{E}$. 
Each internal node $n \in \mathcal{N}$ corresponds to a trainable prototype $\bm{p}_n \in \bm{P}$. 
We follow the prototype definition of ProtoPNet~\cite{NIPS2019_protopnet} where each prototype is a trainable tensor of shape $H_1 \times W_1 \times D$ (with $H_1 \leq H$, $W_1 \leq W$, and in our implementation $H_1=W_1=1$) such that the prototype's depth corresponds to the depth of the convolutional output $\bm{z}$. 

We use a form of generalized convolution without bias~\cite{ghiasi2019generalizing}, where each prototype $\bm{p}_n \in \bm{P}$ acts as a kernel by `sliding' over $\bm{z}$ of shape $H \times W \times D$ and computes the Euclidean distance between $\bm{p}_n$ and its current receptive field $\tilde{\bm{z}}$ (called a \emph{patch}). We apply a minimum pooling operation to select the patch in $\bm{z}$ of shape $H_1 \times W_1 \times D$ that is closest to prototype $\bm{p}_n$:
\begin{equation}
    \tilde{\bm{z}}^* = \argmin_{\tilde{\bm{z}} \in \mathit{patches}(\bm{z})} ||\tilde{\bm{z}} - \bm{p}_n||.
\end{equation}

The distance between the nearest latent patch $\tilde{\bm{z}}^*$ and prototype $\bm{p}_n$ determines to what extent the prototype is present \emph{anywhere} in the input image, which influences the routing of $\bm{z}$ through corresponding node $n$. In contrast to traditional decision trees, where an internal node routes sample $\bm{z}$ either right or left, our node $n \in \mathcal{N}$ is \emph{soft} and routes $\bm{z}$ to both children, each with a fuzzy weight within $[0,1]$, giving it a probabilistic interpretation \cite{frosst2017distilling,irsoy2012soft,Kontschieder_2015_ICCV,ANTS_tanno19a}. 
Following this probabilistic terminology, we define the similarity between $\tilde{\bm{z}}^*$ and $\bm{p}_n$, and therefore the probability of routing sample $\bm{z}$ through the right edge as
\begin{equation}
\label{eq:prob_right}
   p_{e(n, n.\mathit{right})} (\bm{z}) = \exp (- ||\tilde{\bm{z}}^* - \bm{p}_n||),
\end{equation}
such that $p_{e(n, n.\mathit{left})} = 1 - p_{e(n, n.\mathit{right})}$.
Thus, the similarity between prototype $\bm{p}_n$ and the nearest patch in the convolutional output, $\tilde{\bm{z}}^*$, determines to what extent $\bm{z}$ is routed to the right child of node $n$.
Because of the soft routing, $\bm{z}$ is traversed through all edges and ends up in each leaf node $\ell \in \mathcal{L}$ with a certain probability. Path $\mathcal{P}_\ell$ denotes the sequence of edges from the root node to leaf $\ell$. The probability of sample $\bm{z}$ arriving in leaf $\ell$, denoted as $\bm{\pi}_\ell$, is the product of probabilities of the edges in path $\mathcal{P}_\ell$:
\begin{equation}
    \bm{\pi}_\ell(\bm{z}) = \prod_{e \in \mathcal{P}_\ell} p_e(\bm{z}).
\end{equation}

Each leaf node $\ell \in \mathcal{L}$ carries a trainable parameter $\bm{c}_\ell$, denoting the distribution in that leaf over the $K$ classes that needs to be learned. The softmax function $\sigma(\bm{c}_\ell)$ normalizes $\bm{c}_\ell$ to get the class \emph{probability} distribution of leaf $\ell$.
To obtain the final predicted class probability distribution $\bm{\hat{y}}$ for input image $\bm{x}$, 
latent representation $\bm{z} = f(\bm{x}|\bm{\omega})$ is traversed through all edges in $T$ such that all leaves contribute to the final prediction $\bm{\hat{y}}$. An example prediction is shown on the right of Fig.~\ref{fig:prototree_reasoning_overview}. The contribution of leaf $\ell$ is weighted by path probability $\bm{\pi}_\ell$, such that: 
\begin{equation}
    \bm{\hat{y}} (\bm{x}) = \sum_{\ell \in \mathcal{L}} \sigma(\bm{c}_\ell) \cdot \bm{\pi}_\ell(f(\bm{x};\bm{\omega})).
\end{equation}

\section{Training a \modelname}
\label{sec:training}
Training a \modelname requires to learn the parameters $\bm{\omega}$ of CNN $f$ for informative feature maps, the nodes' prototypes $\bm{P}$ for routing and the leaves' class distribution logits $\bm{c}$ for prediction.
The number of prototypes to be learned, \ie $|\bm{P}|$, depends on the tree size. A binary tree structure is initialized by defining a maximum height $h$, which creates $2^{h}$ leaves and $2^{h} - 1$ prototypes. Thus, the computational complexity of learning $\bm{P}$ is growing exponentially with $h$. 

We require a pre-trained CNN $f$ (\eg on ImageNet or training it on a specific prediction task first). During training, prototypes in $\bm{P}$ are trainable tensors. 
Parameters $\bm{\omega}$ and $\bm{P}$ are simultaneously learned with back-propagation by minimizing the cross-entropy loss between the predicted class probability distribution $\bm{\hat{y}}$ and ground-truth $\bm{y}$. 
The learned prototypes should be near a latent patch of a training image such that they can be visualized as an image patch to represent a prototypical part (cf. Sec.~\ref{sec:interpretability}).

\textbf{Learning leaves' distributions.}
In a classical decision tree, a leaf label is learned from the samples ending up in that leaf. Since we use a \emph{soft} tree, learning the leaves' distributions is a global learning problem. Although it is possible to learn $\bm{c}$ with back-propagation together with $\bm{\omega}$ and $\bm{P}$, we found that this gives inferior classification results. 
We hypothesize that including $\bm{c}$ in the loss term leads to an overly complex optimization problem. 
Kontschieder \etal~\cite{Kontschieder_2015_ICCV} noted that solely optimizing leaf parameters is a convex optimization problem and proposed a derivative-free strategy.
Translating their approach to our methodology gives the following update scheme for $\bm{c}_\ell$ for all $\ell \in \mathcal{L}$:
\begin{equation}
\label{eq:offline_classdist}
    \bm{c}_\ell^{(t+1)} = \sum_{\bm{x},\bm{y} \in \bm{\mathcal{T}}} (\sigma(\bm{c}_\ell^{(t)}) \odot \bm{y} \odot \bm{\pi}_\ell) \oslash \bm{\hat{y}},
\end{equation}
where $t$ indexes a training epoch, $\odot$ denotes element-wise multiplication and $\oslash$ is element-wise division. The result is a vector of size $K$ representing the class distribution in leaf $\ell$.
This learning scheme is however computationally expensive: at each epoch, first $\bm{c}_\ell^{(t+1)}$ is computed to update the leaves, and then all other parameters are trained by looping through the data again, meaning that $\bm{\hat{y}}$ is computed twice. 
We propose to do this more efficiently and intertwine mini-batch gradient descent optimization for $\omega$ and $\bm{P}$ with a derivative-free update to learn $\bm{c}$, as shown in Alg.~\ref{alg:train_pseudocode}. 
Our algorithm has the advantage that each mini-batch update of $\omega$ and $\bm{P}$ is taken into account for updating $\bm{c}^{(t+1)}$, which aids convergence. Moreover, computing $\bm{\hat{y}}$ only once for each batch roughly halves the training time.

\begin{algorithm}[t]
\SetAlgoLined
\SetAlgoNoEnd
\KwIn{Training set $\bm{\mathcal{T}}$, max height $h$, $nEpochs$}
 initialize \modelname $T$ with height $h$ and $\bm{\omega}$, $\bm{P}$, $\bm{c}^{(1)}$\;
 \For{$t \in \{1,...,nEpochs\}$}{
  randomly split $\bm{\mathcal{T}}$ into $B$ mini-batches\;
  \For{$(\bm{x}_b,\bm{y}_b) \in \{\bm{\mathcal{T}}_1,...,\bm{\mathcal{T}}_b, ..., \bm{\mathcal{T}}_B\}$}{
  $\bm{\hat{y}}_b = T(\bm{x}_b)$\;
  compute loss $(\bm{\hat{y}}_b, \bm{y}_b)$\;
   update $\omega$ and $\bm{P}$ with gradient descent\;
  \For{$\ell \in \mathcal{L}$}{
   $\bm{c}_\ell^{(t+1)}$ -= $\frac{1}{B} \cdot \bm{c}_\ell^{(t)}$\;
$\bm{c}_\ell^{(t+1)}$ += Eq.~\ref{eq:offline_classdist} for $\bm{x}_b,\bm{y}_b$\;
   }
  }
 }
 prune $T$ (optional)\;
 replace each prototype $\bm{p}_n \in \bm{P}$ with its nearest latent patch $\tilde{\bm{z}}^*_n$ and visualize\;
 \caption{Training a \modelname}
 \label{alg:train_pseudocode}
\end{algorithm}

\section{Interpretability and Visualization}
\label{sec:interpretability}
To foster global model interpretability, we prune ineffective prototypes, visualize the learned latent prototypes, and convert soft to hard decisions.

\subsection{Pruning}
\begin{figure}[t]
\centering
\includegraphics[width=0.9\linewidth]{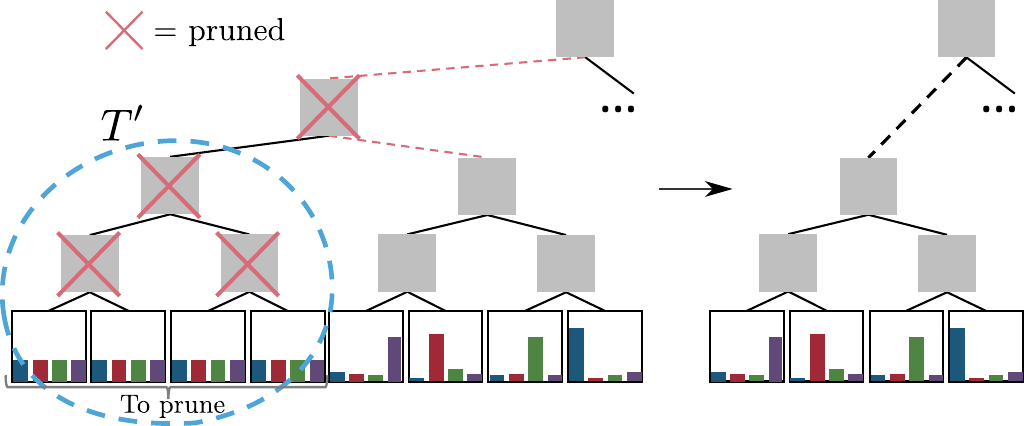}
   \caption{Pruning removes a subtree $T'$, and its parent, in which all leaves have an (nearly) uniform distribution. }
\label{fig:prototree_pruning}
\end{figure}

Interpretability can be quantified by explanation size~\cite{doshi2017towards, silva_towards_2018}. In a \modelname $T$, explanation size is related to the number of prototypes.
To reduce explanation size, 
we analyse the learned class probability distributions in the leaves and remove leaves with nearly uniform distributions, \ie little discriminative power. 
Specifically, we define a threshold $\tau$ and prune all leaves where $\max(\sigma(\bm{c}_\ell)) \leq \tau$,
with $\tau$ being slightly greater than $1/K$ where $K$ is the number of classes.
If all leaves in a full subtree $T' \subset T$ are pruned, $T'$ (and its prototypes) can be removed.
As visualized in Fig.~\ref{fig:prototree_pruning}, \modelname $T$ can be reorganized by additionally removing the now superfluous parent of the root of $T'$.

\subsection{Prototype Visualization}
\label{sec:interpretability_prototype_visualization}
Learned latent prototypes need to be mapped to pixel space to enable interpretability. 
Similar to ProtoPNet~\cite{NIPS2019_protopnet}, we replace each prototype $\bm{p}_n \in \bm{P}$ with its nearest latent patch present in the training data, $\tilde{\bm{z}}^*_n$. 
Thus, 
\begin{equation}
    \bm{p}_n \leftarrow \tilde{\bm{z}}^*_n, \quad \tilde{\bm{z}}^*_n = \argmin_{\bm{z} \in \{f(\bm{x}), \forall \bm{x} \in \bm{\mathcal{T}}\}} ||\tilde{\bm{z}}^* - \bm{p}_n||
    \label{eq:nearest_patch_prototype}
\end{equation}
such that prototype $\bm{p}_n$ is equal to latent representation $\tilde{\bm{z}}^*_n$. Where ProtoPNet replaces its prototypes \emph{during} training every $10^{th}$ epoch, prototype replacement \emph{after} training is sufficient for a \modelname, since our routing mechanism implicitly optimizes prototypes to represent a certain patch. This reduces computational complexity and simplifies the training process.

\begin{figure}[t]
\centering
\includegraphics[width=0.99\linewidth]{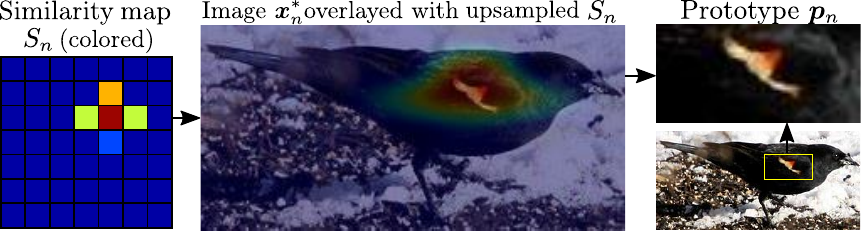}
   \caption{Visualizing a prototype by selecting the most similar patch from the upsampled similarity map.}
\label{fig:prototype_similaritymap}
\end{figure}

We denote by $\bm{x}^*_n$ the training image corresponding to nearest patch $\tilde{\bm{z}}^*_n$.
Prototype $\bm{p}_n$ can now be visualized as a patch of $\bm{x}^*_n$.
We forward $\bm{x}^*_n$ through $f$ to create a 2-dimensional \emph{similarity map} that includes the similarity score between $\bm{p}_n$ and all patches in $\bm{z} = f(\bm{x}^*_n)$
\begin{equation}
    S_n^{(i,j)} = \exp (- ||\tilde{\bm{z}}^{(i,j)} - \bm{p}_n||),
\end{equation}
where $(i,j)$ indicates the location of patch $\tilde{\bm{z}}$ in $\mathit{patches}(\bm{z})$. 
Similarity map $S_n$ is upsampled with bicubic interpolation to the input shape of $\bm{x}^*_n$, after which $\bm{p}_n$ is visualized as a rectangular patch of $\bm{x}^*_n$, at the same location of nearest latent patch $\tilde{\bm{z}}^*_n$ (see Fig.~\ref{fig:prototype_similaritymap}).
Thus, instead of merely showing the nearest training patch in the tree, 
we use the corresponding latent patch $\tilde{\bm{z}}^*_n$ for routing,
making the visualized \modelname a faithful model explanation.

\subsection{Deterministic reasoning}
\label{sec:deterministic_reasoning}
In a soft decision tree, all nodes contribute to a prediction. In contrast, in a hard, deterministic tree, only the nodes along a path account for the final prediction, making hard decision trees easier to interpret than soft trees~\cite{alaniz2019explainable}. Whereas a \modelname is soft during training, we propose two possible strategies to convert a \modelname to a hard tree at test time:
\begin{compactenum}
    \item select the path to the leaf with the highest path probability: $\argmax_{\ell \in \mathcal{L}} (\bm{\pi}_\ell)$
    \item greedily traverse the tree, \ie go right at internal node $n$ if $p_{e(n, n.\mathit{right})} > 0.5$ and left otherwise.
\end{compactenum}
Sec.~\ref{sec:experiments_tradeoff} evaluates to what extent these deterministic strategies influence accuracy compared to soft reasoning.

\section{Experiments}
\label{sec:experiments}
We evaluate the accuracy-interpretability trade-off of a \modelname, and compare our \modelnames with ProtoPNet~\cite{NIPS2019_protopnet} and state-of-the-art models in terms of accuracy and interpretability. 
We evaluate on CUB-200-2011~\cite{WahCUB_200_2011} with 200 bird species (CUB) and Stanford Cars~\cite{cars_dataset_paper} with 196 car types (CARS), since both were used by ProtoPNet~\cite{NIPS2019_protopnet}. 

\subsection{Experimental Setup}
We implemented \modelname in PyTorch. Our CNN $f$ contains the convolutional layers of ResNet50~\cite{he2016deepresnet}, pretrained on ImageNet~\cite{deng2009imagenet} for CARS. For CUB, ResNet50 is pretrained on iNaturalist2017~\cite{Horn_2018_CVPR}, containing plants and animals and therefore a suitable source domain~\cite{li2020rethinking}, using the backbone of~\cite{zhou2020bbn}. Backbone $f$ is frozen for 30 epochs after which $f$ is optimized jointly with the prototypes with Adam~\cite{kingma2014adam}. 
For a fair comparison with ProtoPNet~\cite{NIPS2019_protopnet}, we resize all images to $224 \times 224$ such that the resulting feature maps are $7 \times 7$.
The CNN architecture is extended with a $1 \times 1$ convolutional layer\footnote{ProtoPNet~\cite{NIPS2019_protopnet} appends \emph{two} $1 \times 1$ convolutional layers, but in our model this gave varying (and lower) accuracy across runs.} to reduce the dimensionality of latent output $\bm{z}$ to $D$, the prototype depth. Based on cross-validation from $\{128,256,512\}$, we used $D$=256 for CUB and $D$=128 for CARS. Similar to ProtoPNet, $H_1$=$W_1$=$1$ to provide well-sized patches, such that a prototype is of size $1 \times 1 \times 256$ for CUB.
We use ReLU as activation function, except for the last layer which has a Sigmoid function to act as a form of normalization. 
We initialize the prototypes by sampling from $\mathcal{N}(0.5,0.1)$. 
The initial leaf distributions $\sigma(\bm{c}_\ell^{(1)})$ are uniformly distributed by initializing $\bm{c}_\ell^{(1)}$ with zeros for all $\ell \in \mathcal{L}$. See Suppl. for all details.

\subsection{Accuracy and Interpretability}
\label{sec:experiments_tradeoff}

\begin{table}[t]
    \centering
    \begin{tabular}{>{\centering\arraybackslash}m{0.34cm}>{\centering\arraybackslash}m{3.3cm}>{\centering\arraybackslash}m{0.4cm}>{\arraybackslash}m{1.45cm}>{\centering\arraybackslash}m{0.7cm}}
    \toprule
        Data\newline set & \multicolumn{2}{>{\raggedleft\arraybackslash}m{3.71cm}}{Method \hspace{1.0cm} Inter-\hspace{2.5cm} pret.} & Top-1\newline Accuracy & \#Proto types\\
        \midrule
        \multirow{9}{0.35cm}{\rotatebox[origin=c]{90}{CUB ($224 \times 224$)}} & Triplet Model~\cite{triplet_liang} & -& \textbf{87.5} & \small{n.a.}\\
         & TranSlider~\cite{translider_zhong} & -& $85.8$ & \small{n.a.}\\
         & TASN~\cite{Zheng_2019_CVPR} & o& $87.0$ & \small{n.a.}\\
         \cline{2-5}
         \noalign{\vskip 2pt} 
          & ProtoPNet~\cite{NIPS2019_protopnet} & +& $79.2$ & $2000$\\
         & \textbf{\modelname} $h$=9 (ours) & ++ & $82.2$\small{$\pm 0.7$} & $\bm{202}$\\
         &ProtoPNet ens. (3)~\cite{NIPS2019_protopnet} & +& $84.8$ & $6000$\\
         & \textbf{\modelname} ens. (3) & + & $86.6$ & $605$\\
         &\textbf{\modelname} ens. (5) & + & $\bm{87.2}$ & $1008$\\
        \midrule
        \multirow{8}{0.35cm}{\rotatebox[origin=c]{90}{CARS ($224 \times 224$)}} & RAU~\cite{9219660} & - & \textbf{93.8} & \small{n.a.}\\
        & Triplet Model~\cite{triplet_liang} & -& $93.6$ & \small{n.a.}\\
        & TASN~\cite{Zheng_2019_CVPR} & o & $93.8$ & \small{n.a.}\\
        \cline{2-5}
        \noalign{\vskip 2pt} 
        & ProtoPNet~\cite{NIPS2019_protopnet} & + & $86.1$ & $1960$\\
        & \textbf{\modelname} $h$=11 (ours) & ++ & $86.6$\small{$\pm 0.2$} & $\bm{195}$\\
        &ProtoPNet ens. (3)~\cite{NIPS2019_protopnet} & + & $91.4$ & $5880$\\
        & \textbf{\modelname} ens. (3) & + & $90.3$ & $586$\\
        &\textbf{\modelname} ens. (5) & + & $\bm{91.5}$ & $977$\\
        \bottomrule
    \end{tabular}
    \caption[]{Mean accuracy and standard deviation of our \modelname (5 runs) and ensemble with 3 or 5 \modelnames compared with self-reported accuracy of uninterpretable state-of-the-art\protect\footnotemark~(-), attention-based models (o) and interpretable ProtoPNet (+, with ResNet34-backbone).}
    \label{tab:accuracies}
\end{table}
\footnotetext{Using higher-resolution images (\eg $448 \times 448$) has shown to give better results~\cite{NIPS2019_9035,Zheng_2019_CVPR} with \eg accuracy up to 90.4\%~\cite{Ge_2019_CVPR} for CUB.}

\begin{figure}[tb]
\begin{center}
\includegraphics[width=0.96\linewidth]{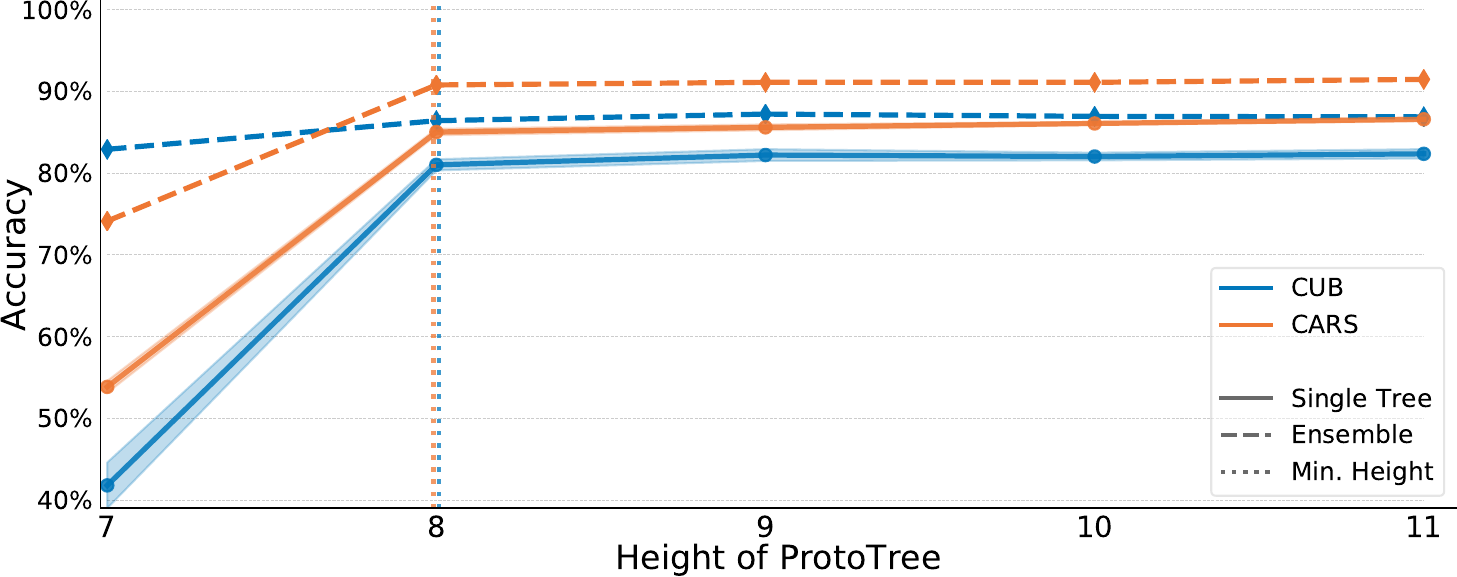}
\end{center}
\caption{Top-1 accuracy of a \modelname (across 5 runs), and an ensemble with those 5 \modelnames. A vertical dotted line shows the minimal height such that \#leaves $\geq$ \#classes.}
\label{fig:accuracy_height}
\end{figure}

Table~\ref{tab:accuracies} compares the accuracy of \modelnames with state-of-the-art methods. Our \modelname outperforms ProtoPNet for both datasets. We also evaluated the accuracy of \modelname ensembles by averaging the predictions of 3 or 5 individual \modelnames, all trained on the same dataset. An ensemble of \modelnames outperforms a ProtoPNet ensemble, and approximates the accuracy of uninterpretable or attention-based methods, while providing intrinsically interpretable global and faithful local explanations.

\begin{table*}
    \centering
    \begin{tabular}{>{\centering\arraybackslash}m{0.8cm} >{\centering\arraybackslash}m{0.35cm} >{\centering\arraybackslash}m{0.25cm} | c c c c c c}
        \toprule
        Dataset & $K$ & $h$ & Initial Acc & Acc pruned & Acc pruned+repl. & \# Prototypes & \% Pruned & Distance $\tilde{\bm{z}}^*_n$\\
        \midrule
        CUB & 200 & 9 & $82.206 \pm 0.723$ & $82.192 \pm 0.723$ & $82.199 \pm 0.726$ & $201.6 \pm 1.9$ & 60.5 & $0.0020 \pm 0.0068$\\
        CARS & 196 & 11 & $86.584 \pm 0.250$ & $86.576 \pm 0.245$ & $86.576 \pm 0.245$ & $195.4 \pm 0.5$ & 90.5 & $0.0005 \pm 0.0016$ \\
        \bottomrule
    \end{tabular}
    \caption{Impact of pruning and prototype replacement: 1) before pruning and replacement, 2) after pruning, 3) after pruning and replacement, 4) number of prototypes after pruning, 5) fraction of prototypes that is pruned and 6) Euclidean distance from each latent prototype to its nearest latent training patch (after pruning). Showing mean and std dev across 5 runs.}
    \label{tab:pruning_projection_acc}
\end{table*}

\textbf{Tree Height.}
A \modelname with fewer prototypes is smaller and hence easier to interpret, but represents a less complex model and might have less predictive power. 
Fig.~\ref{fig:accuracy_height} shows the accuracy of \modelnames with increasing height.
It confirms that it is sensible to set the initial height $h$ such that the number of leaves is at least as large as the number of classes $K$.  For CUB, accuracy increases up to a certain height ($h=9$) after which accuracy plateaus. 
An increasing height has a higher impact on the accuracy for CARS, probably because of its lower inter-class part similarity for which a more imbalanced tree, with fewer shared prototypes, is more suitable.  
Ensembling substantially increases prediction accuracy, although at the cost of explanation size. 

\begin{figure*}
    \centering
    \includegraphics[width=0.812\textwidth]{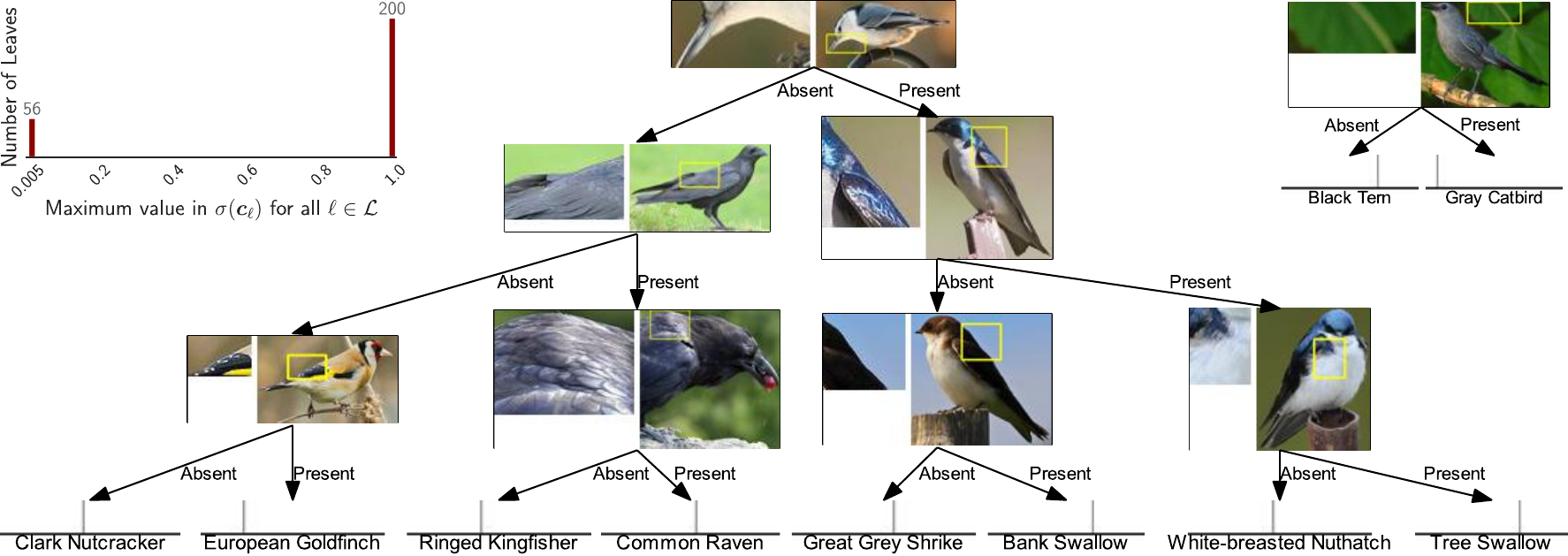}
    \caption{Subtree of an automatically visualized \modelname trained on CUB, $h$=8 (middle). 
    Each internal node contains a prototype (left) and the training image from which it is extracted (right). A \modelname faithfully shows its reasoning and clusters similar classes (\eg birds with a white chest). Top left: maximum values of all leaf distributions. Top right: \modelname reveals biases learned by the model: \eg classifying a Gray Catbird based on the presence of a leaf. Best viewed in color.}
    \label{fig:prototree_snippet}
\end{figure*}

\textbf{Pruning.}  
Since our training algorithm optimizes the leaf distributions to minimize the error between $\bm{\hat{y}}$ and one-hot encoded $\bm{y}$, most leaves learn either one class label, or an almost uniform distribution, as shown in Fig.~\ref{fig:prototree_snippet} (top left) for CUB with $h$=8.
Other datasets and tree heights show a similar pattern (Suppl.). We set pruning threshold $\tau=0.01$, such that we are left with leaves that can be interpreted (nearly) deterministically. Table~\ref{tab:pruning_projection_acc} shows that the prediction accuracy of a \modelname barely changes when the tree is pruned and visualized. The negligible difference after prototype replacement (\ie visualization) is supported by the fact that the distance from each prototype to its nearest patch is close to zero, indicating that \modelname already implicitly optimizes prototypes to be near a latent image patch. This confirms that replacing prototypes after training suffices, instead of \emph{during} training as in ProtoPNet~\cite{NIPS2019_protopnet}.
Furthermore, pruning drastically reduces the size of the tree (up to $>90\%$), preserving roughly 1 prototype per class. 
In contrast, ProtoPNet~\cite{NIPS2019_protopnet} uses 10 prototypes per class (cf. Tab.~\ref{tab:accuracies}), resulting in 2000 prototypes in total for CUB.  Thus, a \modelname is almost 90\% smaller and therefore easier to interpret. Even with an ensemble of \modelnames, which increases global explanation size, the number of prototypes is still substantially smaller than ProtoPNet (cf.~Table~\ref{tab:accuracies}).

\textbf{Deterministic reasoning.}
\begin{table}
    \centering
    \begin{tabular}{>{\centering\arraybackslash}m{1.10cm} >{\centering\arraybackslash}m{1.70cm} >{\centering\arraybackslash}m{1.88cm} >{\centering\arraybackslash}m{2.14cm}}
    \toprule
        Strategy & Accuracy & Fidelity & Path length \\
        \midrule
        Soft &  $82.20 \pm 0.01$ & n.a. & n.a. \\
        Max $\bm{\pi}_\ell$ &  $82.19 \pm 0.01$ & $0.999 \pm 0.001$ & $8.3 \pm 1.1$ $(9,3)$ \\
        Greedy & $82.07 \pm 0.01$ & $0.987 \pm 0.002$ & $8.3 \pm 1.1$ $(9,3)$\\
        \bottomrule
    \end{tabular}
    \caption{
    Soft vs. deterministic classification strategies at test time.  Fidelity is agreement with soft strategy. Min and max path lengths in brackets.
ProtoTree on CUB ($h$=9, pruned and replaced), averaged over 5 runs (mean, stdev).}
    \label{tab:deterministic}
\end{table}
\begin{figure*}[ht]
    \centering
    \includegraphics[width=0.96\textwidth]{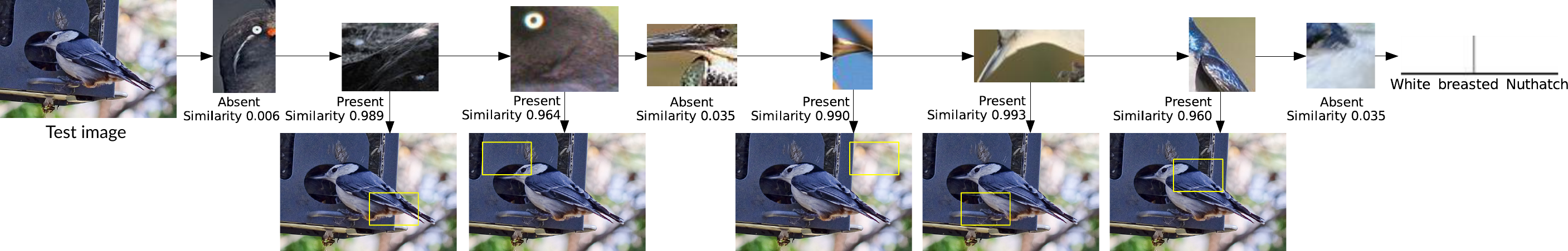}
    \caption{Local explanation that shows the greedy path when classifying a White-breasted Nuthatch. Prototypes found in the test image are: a dark-colored tail, a contrastive eye, sky background (learned bias), a white chest and a blue-grey wing.}
    \label{fig:prototree_local_explanation}
\end{figure*}
As discussed in Sec.~\ref{sec:deterministic_reasoning}, a \modelname can make deterministic predictions at test time to improve human understanding of the decision making process. Table~\ref{tab:deterministic} shows that
selecting the leaf with the highest path probability leads to nearly the same prediction accuracy as soft routing, since the fidelity (\ie fraction of test images for which the soft and hard strategy make the same classification~\cite{guidotti2018survey}) is practically 1. The greedy strategy performs slightly worse but its fidelity is still close to 1. Results are similar for other datasets and tree heights (Suppl.). Our experiments therefore show that a \modelname can be safely converted to a deterministic tree, such that a prediction can be explained by presenting one path in the tree. Compared to ProtoPNet~\cite{NIPS2019_protopnet}, where a user is required to analyse 2000 prototypes to understand a single prediction for CUB, our deterministic \modelname ($h$=9) reduces the number of decisions to follow to 9 prototypes at maximum. When using a more accurate ensemble of 5 deterministic \modelnames, a maximum of only 45 prototypes needs to be analysed, resulting in much smaller local explanations than ProtoPNet.

\textbf{Visualizations and Discussion.}
Figure~\ref{fig:prototree_snippet} shows a snippet of a \modelname trained on CUB (more in Suppl.), and Figure~\ref{fig:prototree_local_explanation} shows a local explanation containing the full path along the tree with a greedy classification strategy. From analysing various \modelnames, we conclude that prototypes are in general perceptually relevant, and successfully cluster similar-looking classes. Similar to ProtoPNet~\cite{NIPS2019_protopnet}, some prototypes seem to focus on background. This is not necessarily an error in our visualization but shows that a \modelname can reveal 
learned biases. For example, Fig.~\ref{fig:prototree_snippet} (top right) shows a green leaf to distinguish between a Gray Catbird and a Black Tern, because the latter is in the training data usually surrounded by sky or water. 
Further research could investigate to what extent undesired prototypes can be `fixed' with a human-in-the-loop that replaces them with a manually selected patch, in order to create a model that is completely ``right for the right reasons''~\cite{ross2017right}.
Furthermore, we found that human's perceptual similarity could differ from similarity assigned by the model, since it is not always clear why the model considered an image highly similar to a prototype. The visualized prototypes could therefore be further explained by indicating whether \eg color or shape was most important, as presented by~\cite{nauta2020looks}, or by showing a cluster of patches. Especially prototypes close to the root of the tree are sometimes not as clear and semantically meaningful as prototypes closer to leaves. This is probably due to the binary tree structure that requires a patch from a training image to split the data into two subsets. A natural progression of this work would therefore be to investigate non-binary trees, with multiple prototypes per node. 

\section{Conclusion}
We presented the \fullmodelname (\modelname) for intrinsically interpretable fine-grained image recognition. 
Whereas the Prototypical Part Network (ProtoPNet)~\cite{NIPS2019_protopnet} presents a user a large number of prototypes, our novel architecture with end-to-end training procedure improves interpretability by arranging the prototypes in a hierarchical tree structure. This breaks up the reasoning process in small steps which simplifies model comprehension and error analysis, and reduces the number of prototypes by a factor of 10.
Most learned prototypes are semantically relevant, which results in a fully simulatable model. 
Additionally, we outperform ProtoPNet~\cite{NIPS2019_protopnet} on the CUB-200-2011 and Stanford Cars data sets. An ensemble of 5 \modelnames approximates the accuracy of non-interpretable state-of-the art models,
while still having fewer prototypes than ProtoPNet~\cite{NIPS2019_protopnet}. 
Thus, \modelname achieves competitive performance while maintaining intrinsic interpretability. As a result, our work questions the existence of an accuracy-interpretability trade-off and stimulates novel usage of powerful neural networks as backbone for interpretable, predictive models.
In future work, we would like to investigate the potential of ProtoTree for other types of problems that contain prototypical features, such as specific wave patterns in sensor data. 
\newpage
{\small
\bibliographystyle{ieee_fullname}
\bibliography{main_bib}
}

\end{document}


\title{Neural Prototype Trees for Interpretable Fine-grained Image Recognition: Supplementary Material}

\author{Meike Nauta\textsuperscript{1} \quad Ron van Bree\textsuperscript{1} \quad Christin Seifert\textsuperscript{1,2}\\
\textsuperscript{1} University of Twente, the Netherlands \quad
\textsuperscript{2} University of Duisburg-Essen, Germany\\
{\tt\small m.nauta@utwente.nl}, {\tt\small r.j.vanbree@student.utwente.nl}, {\tt\small christin.seifert@uni-due.de}
}

\maketitle

\setcounter{page}{1}
\makeatletter
\renewcommand{\theequation}{S\arabic{equation}}
\renewcommand{\thetable}{S\arabic{table}}
\renewcommand{\thefigure}{S\arabic{figure}}
\renewcommand{\thesection}{S\arabic{section}}

\section{Training Details}
\label{sec:supp_details}
We train the neural network $f$ and the prototypes of a \modelname with Adam~\cite{kingma2014adam}, and the leaves with our derivative-free algorithm. As shown in Table~\ref{tab:network_architecture}, the prototypes and leaves are only a fraction of the trainable parameters and therefore barely give any overhead. However, note that the number of prototypes and number of leaves will exponentially increase when increasing height $h$. 

\begin{table}[ht!]
    \centering
    \begin{tabular}{p{0.25cm} p{2.45cm}|p{2.3cm} p{1.55cm}}
    \toprule
        Part & Layer &  (Output) Shape & Total \# Parameters\\
        \midrule
        \multirow{3}{*}{$f$} & ResNet50 \footnotesize{(without avgpool and fc-layer)} & (2048, 7, 7) & 23,508,032\\
        & $1\times1$ Conv2D & (256, 7, 7) & 524,288\\
        $\bm{P}$ & & $255 \times 1 \times 1 \times 256$ & 65,280\\
        $\bm{c}$ & & $256 \times 200$ & 51,200 \\
        \midrule
        Total & & & 24M\\
        \bottomrule
    \end{tabular}
    \caption{Trainable parameters in a \modelname with height $h = 8$ and $D = 256$, for CUB-200-2011 with 200 classes.}
    \label{tab:network_architecture}
\end{table}

For CUB, we use the backbone of~\cite{zhou2020bbn} pretrained on iNaturalist2017 for 180 epochs. For CARS, we use a ResNet50 pretrained on ImageNet. This backbone, except for the last convolutional layer, is frozen for some epochs (Table~\ref{tab:parameter_values}). 
The $1 \times 1$ convolutional layer is initialized with Xavier initialization~\cite{glorot2010understanding}. The prototypes, the last layers of $f$, and the backbone each have their specified learning rate, as indicated in Table~\ref{tab:parameter_values}.

\begin{table}[ht]
    \centering
    \begin{tabular}{p{0.75cm} p{4.1cm}|p{2.2cm}}
    \toprule
        Data & Parameter &  Value\\
        \midrule
        
        \multirow{10}{0.6cm}{All} 
        & Batch size & 64 \\
        & Input image size & $224 \times 224$ \\
        & Output image size & $7 \times 7$ \\
       & $H_1$ & 1 \\
        & $W_1$ & 1 \\
        & Learning rate prototypes & 0.001\\
        & Learning rate $1 \times 1$ conv layer and last conv layer ResNet50 & 0.001\\
        & Gamma for lr decay & 0.5\\
        & Epochs backbone frozen & 30\\
        \midrule
        \multirow{6}{0.75cm}{CUB} & Learning rate pretrained ResNet50 (except last layer) & $1e-5$\\
        & Pruning threshold $\tau$ & 0.01\\
         & Number of epochs (after offline data augmentation) & 100 \\
         & Milestones for lr decay & 60,70,80,90,100\\
        \midrule
        \multirow{6}{0.75cm}{CARS} & Learning rate pretrained ResNet50 (except last layer) & $2e-4$\\
        & Pruning threshold $\tau$ & 0.01\\
        & Number of epochs & 500 \\
        & Milestones for lr decay & 250,350,400,425, 450,475,500\\
        \bottomrule
    \end{tabular}
    \caption{Parameter values when training \modelnames for our experiments.}
    \label{tab:parameter_values}
\end{table}

\paragraph{Data Augmentation} For CUB, we crop each training image offline into four corners based on the bounding box annotations, and include the full image resulting in 5 images per original image. We then resize each image to $224 \times 224$. Test images are not cropped and resized to $224 \times 224$. To make our visualizations comparable to ProtoPNet~\cite{NIPS2019_protopnet}, we select the nearest training image patch for each prototype by considering cropped training images only. 

For CARS, we do not use any annotations. We resize all images to $256 \times 256$, apply online data augmentation and then take a random crop of size $224 \times 224$. 
Comparable to ProtoPNet~\cite{NIPS2019_protopnet_supp}, we apply data augmentation including random rotation, shearing, distortion, color jitter and horizontal flipping. Data augmentation details (applied in an online fashion and implemented in PyTorch) are shown in Table~\ref{tab:data_augmentation}. More complex training and augmentation techniques, such as AutoAugment~\cite{Cubuk_2019_CVPR} and cyclic learning rates~\cite{lr_finder_cyclic_smith}, are not used to keep a fair comparison, but might improve accuracy. Similarly, applying more advanced ensemble techniques, such as bagging and boosting, might improve the prediction accuracy of a \modelname ensemble. 

\begin{table}[hbt]
    \centering
    \begin{tabular}{p{0.6cm} p{3.05cm}|p{3.4cm}}
    \toprule
        Data & Augmentation &  Value/Scale\\
        \midrule
        \multirow{7}{0.6cm}{All} & 
        Brightness jitter & (0.6, 1.4)\\
        & Contrast jitter & (0.6, 1.4)\\
        & Saturation jitter & (0.6,1.4)\\
        & Horizontal flip & $p=0.5$ \\
        & Random shear & (-2, 2) \\
        & Normalization & mean 0.485,0.456,0.406 std 0.229,0.224,0.225\\
        \midrule
        \multirow{5}{0.6cm}{CUB} & Hue jitter & (-0.02, 0.02)\\
        & Random rotation & 10\\ 
        & Random translation & (0.05, 0.05) \\
        & Perspective distortion & 0.2 ($p=0.5$)\\
        & Resize & ($224 \times 224$) \\
        \midrule
        \multirow{5}{0.6cm}{CARS} & 
        Hue jitter & (-0.4, 0.4)\\
        & Random rotation & 15\\ 
        & Perspective distortion & 0.5 ($p=0.5$)\\
        & Resize & ($256 \times 256$) \\
        & Random Crop & ($224 \times 224$)\\
        \bottomrule
    \end{tabular}
    \caption{Online data augmentation. Jitter values are based on~\cite{he2019bag}, except for smaller hue differences since color hue can be discriminative for classes in CUB.}
    \label{tab:data_augmentation}
\end{table}

\section{Prototype Visualization with Class Constraints}
\label{sec:supp_prototype_vis}

Prototypes are trainable vectors that, after training, can be replaced with a latent patch of a training image.
Equation 6 (main paper) shows that the nearest training patch $\tilde{\bm{z}}^*_n$ can be found by looping through all images in the training set. 
Whereas ProtoPNet has class-specific prototypes, our prototypes can be of any class. However, we argue that the perceptual interpretability of a prototype in \modelname $T$ can be improved by only considering images that have a certain class label. 

Specifically, we require that $\bm{x}^*_n$ should be from the majority class of any of the leaves reachable by node $n$. For each internal node $n$ and corresponding prototype $\bm{p}_n$, we define $T'_n \subset T$ as a full binary subtree of $T$ with $n$ as root node, such that $\mathcal{Y}'_n$ is the corresponding set of class labels $\{\argmax_{\bm{c}_\ell}$ for all $\ell$ in $T'_n$\}. $\bm{\mathcal{T}}'_n \subseteq \bm{\mathcal{T}}$ is the set of training images with class label $\in \mathcal{Y}'$. Then, Equation 6 from the main paper can be adapted as follows:
\begin{equation}
    \bm{p}_n \leftarrow \tilde{\bm{z}}^{*'}_n, \quad \tilde{\bm{z}}^{*'}_n = \argmin_{\bm{z} \in \{f(\bm{x}), \forall \bm{x} \in \bm{\mathcal{T}}'_n\}} ||\tilde{\bm{z}}^* - \bm{p}_n||.
    \label{eq:supp_nearest_patch_prototype_restrictions}
\end{equation}

\begin{figure*}[tbh]
    \centering
    \includegraphics[width=0.82\textwidth]{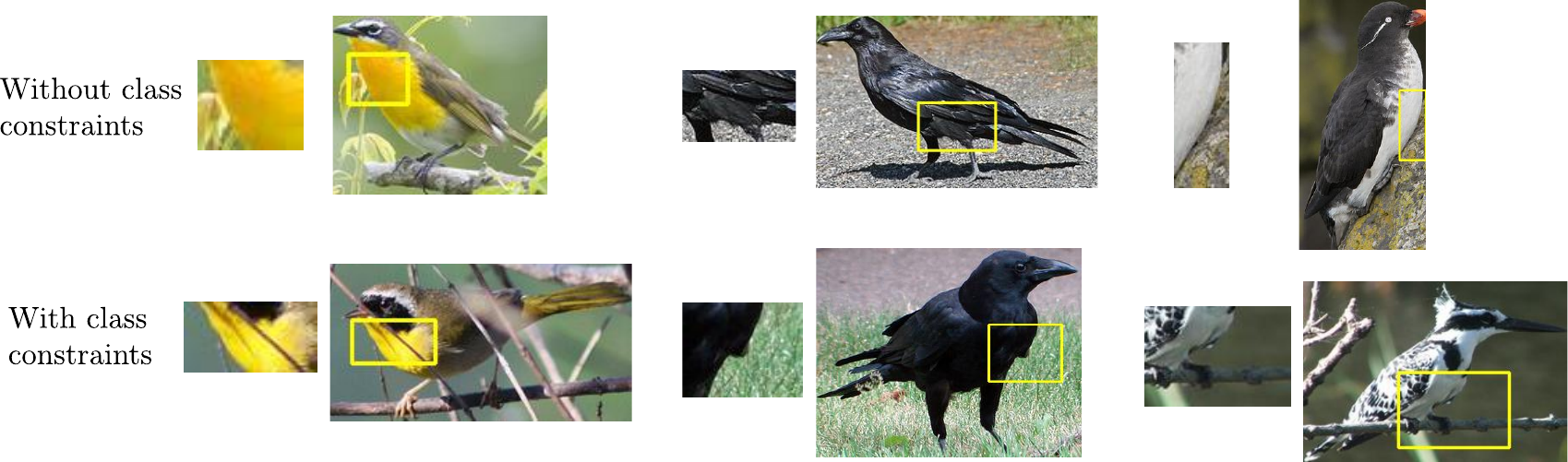}
    \caption{Three prototypes occurring in a \modelname trained on CUB. The upper row shows prototypes when considering all images for prototype replacement (Eq. 6). The bottom row shows prototypes when only those images are considered that have a class label that is from the majority class of any of the reachable leaves (Eq.~\ref{eq:supp_nearest_patch_prototype_restrictions}). For example, the left column shows that the prototype represents a white belly. For a human classifying a bird similar to the bottom left image, perceptual similarity might be higher for the bottom left prototype than the upper left prototype.}
    \label{fig:supp_projection_class_constraints}
\end{figure*}

We denote by $\bm{x}^*_n$ the training image corresponding to nearest patch $\tilde{\bm{z}}^*_n$ when considering all training data, and $\bm{x}^{*'}_n$ denotes the training image corresponding to nearest patch $\tilde{\bm{z}}^*_n$ with class restrictions as defined in Equation~\ref{eq:supp_nearest_patch_prototype_restrictions}.
In our experiments on CUB, we found that 
the difference in Euclidean distance from $\bm{p}_n$ to $\tilde{\bm{z}}^{*'}_n$ with $\bm{p}_n$ to $\tilde{\bm{z}}^{*}_n$ was 
$\num{5.86e-05}$ on average, and is therefore negligible. Figure~\ref{fig:supp_projection_class_constraints} visualizes three prototypes with and without such constraints ($\tilde{\bm{z}}^{*'}_n$ and $\tilde{\bm{z}}^{*}_n$). Both visualization methods (with or without class contraints) also give a similar prediction accuracy, as shown in 
Table~\ref{tab:supp_constrained_visualization_comparison}. Interestingly, adding the class constraints even slightly improves accuracy.

\begin{table}[ht]
    \centering
    \begin{tabular}{c|c}
    \toprule
    Visualization method & Accuracy\\
    \midrule
    Without class constraints (Eq. 6) & $82.195 \pm 0.723$\\
    With class constraints (Eq.~\ref{eq:supp_nearest_patch_prototype_restrictions}) & $82.199 \pm 0.726$\\
    \bottomrule
    \end{tabular}
    \caption{Accuracy of \modelname after pruning and visualization for CUB ($h=9$) across 5 runs.}
    \label{tab:supp_constrained_visualization_comparison}
\end{table}

Thus, adding the restriction that $\bm{x}^*_n$ should be from the majority class of any of the leaves reachable by node $n$ does not negatively impact the accuracy of the model, but could improve interpretability. Our results in the main paper and Supplementary material are based on prototype replacement with class constraints.

\section{Detailed Results}
\label{sec:supp_detailed_results}
Table~\ref{tab:supp_deterministic} compares the deterministic classification strategies with the soft strategy for a ProtoTree trained on CARS. It shows that, similar to the results for CUB, selecting the leaf with the highest path probability leads to nearly the same prediction accuracy as soft routing, since the fidelity is 1. The greedy strategy performs slightly worse but its fidelity is still close to 1. Interestingly, pruning a ProtoTree of height 11 trained on CARS leads to a much smaller tree, with an average path length of only 8.6. 
\begin{table}[!b]
    \centering
    \begin{tabular}{>{\centering\arraybackslash}m{1.10cm} >{\centering\arraybackslash}m{1.35cm} >{\centering\arraybackslash}m{1.88cm} >{\centering\arraybackslash}m{2.31cm}}
    \toprule
        Strategy & Accuracy & Fidelity & Path length \\
        \midrule
        Soft &  $86.58 \pm $\small{$0.24$} & n.a. & n.a. \\
        Max $\bm{\pi}_\ell$ &  $86.58 \pm $\small{$0.24$} & $1.000 \pm 0.000$ & $8.6 \pm 1.7$ $(11,4)$ \\
        Greedy & $86.43 \pm $\small{$0.30$} & $0.992 \pm 0.002$ & $8.6 \pm 1.7$ $(11,4)$\\
        \bottomrule
    \end{tabular}
    \caption{
    Soft vs. deterministic classification strategies at test time.  Fidelity is agreement with soft strategy. Min and max path lengths in brackets.
ProtoTree on CARS ($h$=11, pruned and replaced), averaged over 5 runs (mean, stdev).}
    \label{tab:supp_deterministic}
\end{table}

\begin{figure*}[htb]
    \centering
    \begin{subfigure}[t]{0.31\linewidth}
        \centering
        \includegraphics[width=0.98\textwidth]{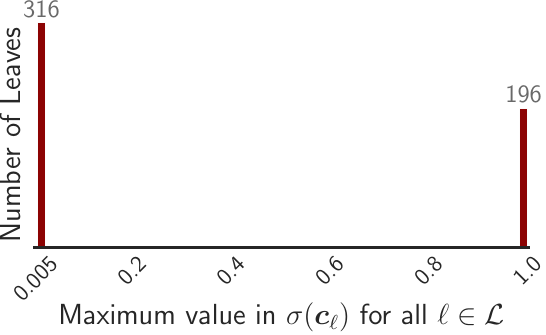}
        \caption{CARS, $h=9$}
        \label{fig:histogram_cars_h9}
    \end{subfigure}\hfill
    \begin{subfigure}[t]{0.31\linewidth}
        \centering
        \includegraphics[width=0.98\textwidth]{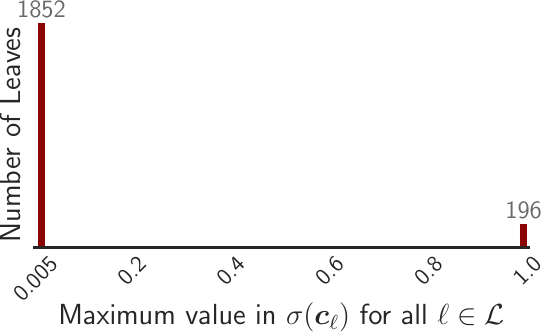}
        \caption{CARS, $h=11$}
        \label{fig:histogram_cars_h11}
    \end{subfigure}\hfill
    \begin{subfigure}[t]{0.31\linewidth}
        \centering
        \includegraphics[width=0.98\textwidth]{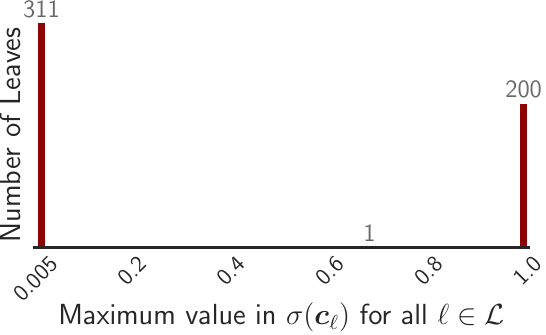}
        \caption{CUB, $h=9$}
        \label{fig:histogram_cub_h9}
    \end{subfigure}
    \caption{Maximum values of all leaf distributions in a trained ProtoTree.}
    \label{fig:supp_histograms}
\end{figure*}

Figure~\ref{fig:supp_histograms} shows the maximum values of all leaf distributions for trained ProtoTrees on CARS or CUB. It can be seen that almost all leaves learn either one class label, or an almost uniform distribution ($1/K$).

Table~\ref{tab:supp_pruning_projection_acc_all}) presents the detailed results for ProtoTrees of various heights trained on CUB or CARS.

\begin{table*}[!b]
    \centering
    \begin{tabular}{>{\centering\arraybackslash}m{1.4cm} c | c c c c c c}
        \toprule
        Dataset & $h$ & Initial Acc & Acc pruned & Acc pruned+vis. & \# Prototypes & \% Pruned & Distance $\tilde{\bm{z}}^*_n$\\
        \midrule
        \multirow{4}{1.6cm}{CUB ($K=200$)} & 7 & $41.826 \pm 2.776$ & $41.826 \pm 2.776$ & $41.798 \pm 2.780$ & $127.0 \pm 0.0$ & 0.0 & $0.0027 \pm 0.0045$ \\
        & 8 & $81.046 \pm 0.674$ & $81.042 \pm 0.676$ & $81.032 \pm 0.680$ & $200.4 \pm 1.2$ & 21.4 & $0.0025 \pm 0.0047$ \\
         & 9 & $82.206 \pm 0.723$ & $82.192 \pm 0.723$ & $82.199 \pm 0.726$ & $201.6 \pm 1.9$ & 60.5 & $0.0020 \pm 0.0068$\\
         & 10 & $82.054 \pm 0.517$ & $82.019 \pm 0.468$ & $82.019 \pm 0.469$& $203.2 \pm 2.0$ & 80.1 & $0.0018 \pm 0.0072$\\
         & 11 & $82.370 \pm 0.575$ & $82.357 \pm 0.580$ & $82.352 \pm 0.572$& $207.0 \pm 2.7$ & 89.9 & $0.0038 \pm 0.0313$\\
          \midrule
          \multirow{4}{1.6cm}{CARS ($K=196$)} & 7 & $53.842 \pm 0.733$ & $53.842 \pm 0.733$ & $53.847 \pm 0.732$ & $127.0 \pm 0.0$ & 0.0 & $0.0006 \pm 0.0018$ \\
        & 8 & $85.049 \pm 0.384$ & $85.007 \pm 0.398$ & $85.017 \pm 0.393$ & $195.0 \pm 0.0$ & 23.5 & $0.0005 \pm 0.0018$ \\
         & 9 & $85.601 \pm 0.361$ & $85.586 \pm 0.361$ & $85.586 \pm 0.361$ & $195.2 \pm 0.4$ & 61.8  & $0.0027 \pm 0.0626$ \\
         & 10 & $86.064 \pm 0.187$ & $86.071 \pm 0.191$ & $86.076 \pm 0.186$ & $195.8 \pm 1.2$ & 80.9 & $0.0005 \pm 0.0017$ \\
         & 11 & $86.584 \pm 0.250$ & $86.576 \pm 0.245$ & $86.576 \pm 0.245$ & $195.4 \pm 0.5$ & 90.5 & $0.0005 \pm 0.0016$ \\
        \bottomrule
    \end{tabular}
    \caption{Mean and standard deviation across 5 runs of: 1) accuracy before pruning and visualization, 2) accuracy after pruning, 3) accuracy after pruning and visualization, 4) number of prototypes after pruning, 5) fraction of prototypes that is pruned and 6) Euclidean distance from each latent prototype to its nearest latent training patch (after pruning).}
    \label{tab:supp_pruning_projection_acc_all}
\end{table*}

\clearpage
\onecolumn
\section{More Visualized ProtoTrees}

\begin{figure*}[!ht]
    \centering
    \includegraphics[width=\textwidth]{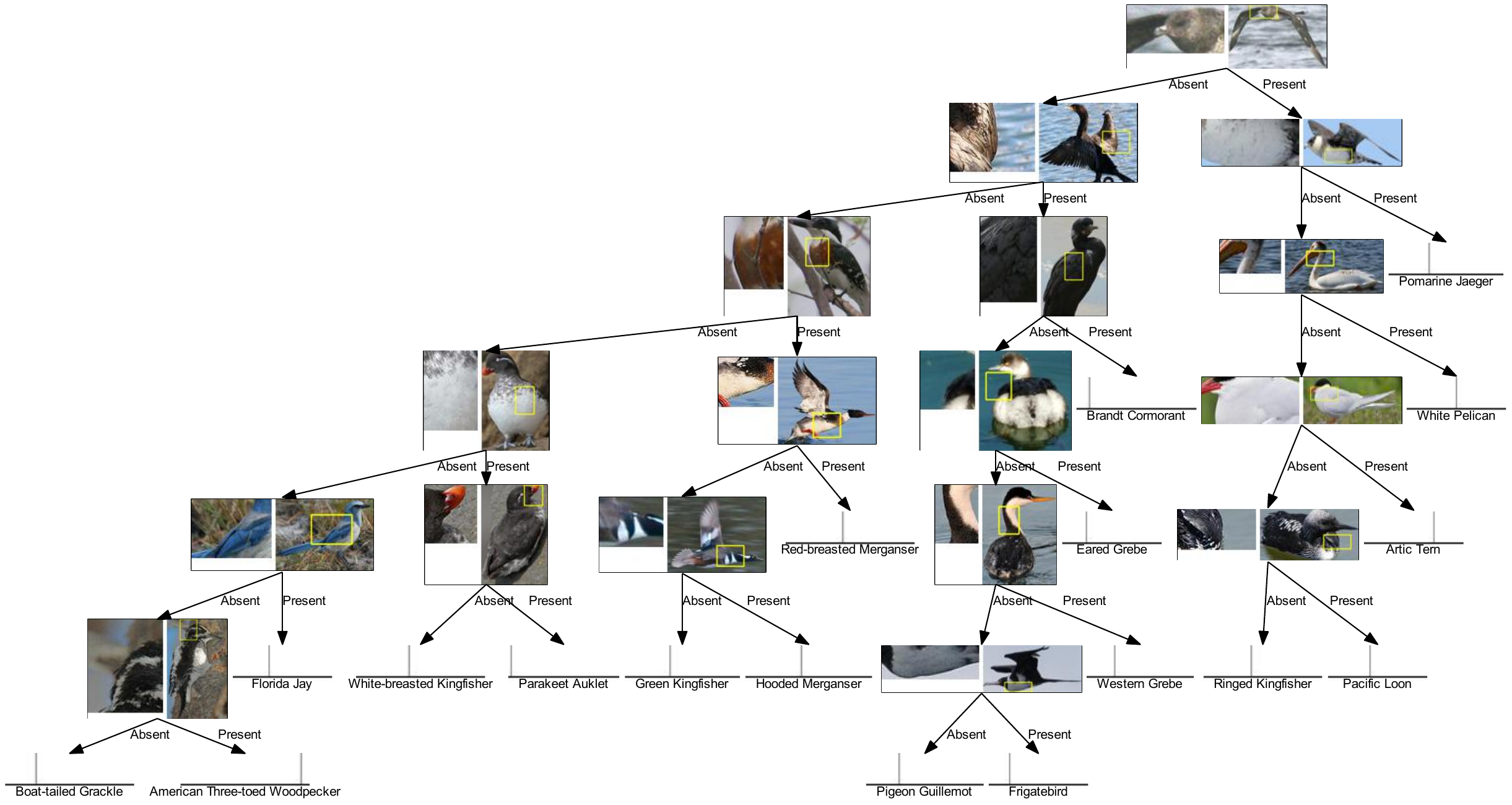}
    \caption{Subtree of a ProtoTree (CUB, $h=9$). Each internal node contains a prototype (left) and the training image from which it is extracted (right). Each leaf shows the class probability distribution and the label of the class with the highest probability. Prototypes seem to correctly represent distinctive parts. For example, the Green Kingfisher, Hooded Merganser and Red-breasted Merganser all have a red-brown spot. Interpreting the top node is a bit challenging. A local explanation showing the similarity with a test image, or supplementary explanations as presented in~\cite{nauta2020looks} could help to clarify this.}
    \label{fig:cub_h9_bigexample}
\end{figure*}
\begin{figure*}[!hb]
    \centering
    \includegraphics[width=\textwidth]{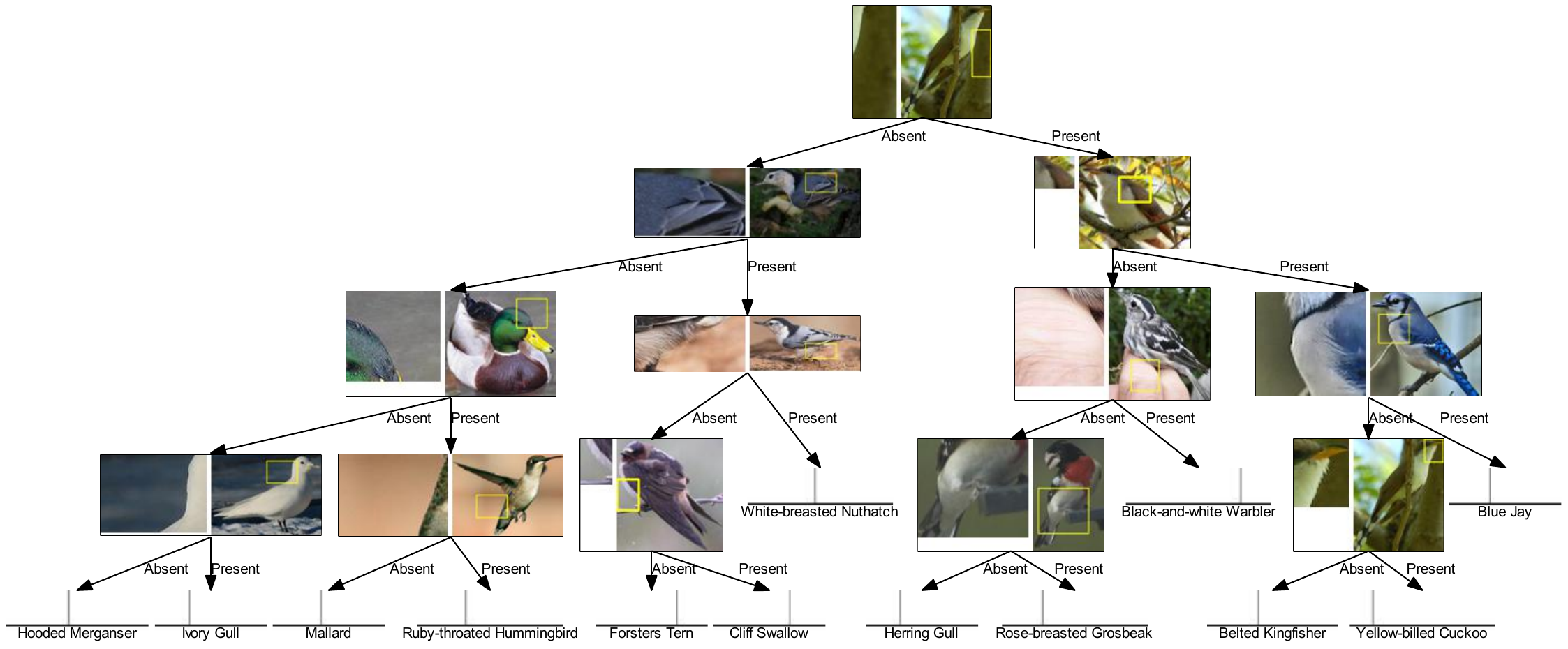}
    \caption{Subtree of an automatically visualized ProtoTree (CUB, $h=8$). Each internal node contains a prototype (left) and the training image from which it is extracted (right). The Mallard and Ruby Throated Hummingbird share the same green-colored prototype.}
    \label{fig:cub_h8_example1}
\end{figure*}

\begin{figure*}
    \centering
    \includegraphics[width=\textwidth]{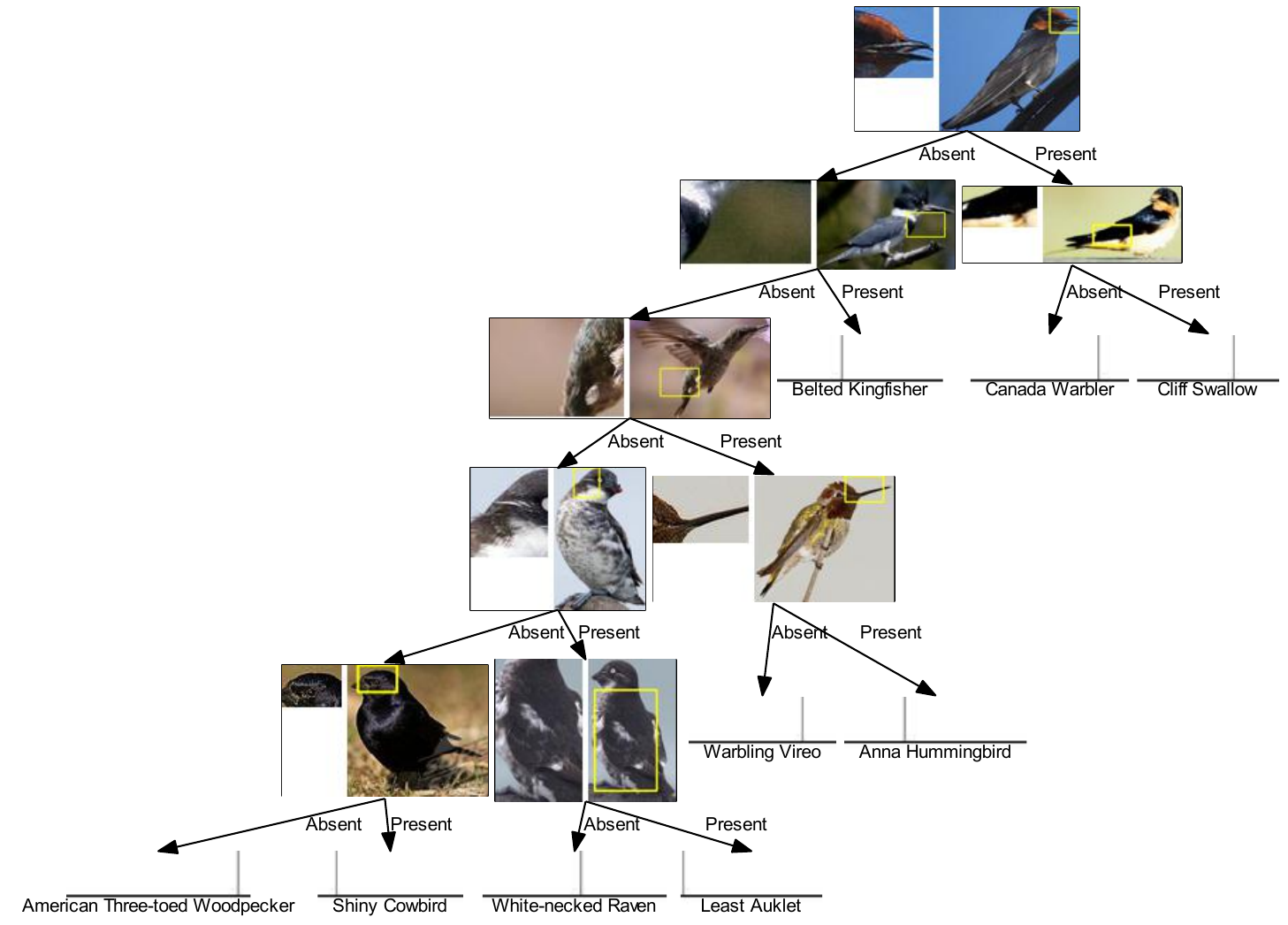}
    \caption{Subtree of a ProtoTree (CUB, $h=10$). The Anna Hummingbird is recognized by its specific, long bill. Generally, a higher maximum height $h$ results, after pruning, in a less balanced tree.}
    \label{fig:cub_h10}
\end{figure*}

\begin{figure*}
    \centering
    \includegraphics[width=\textwidth]{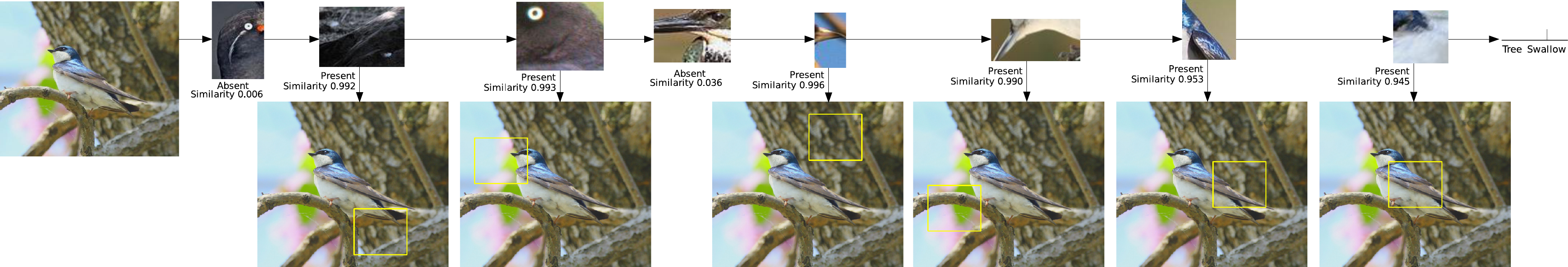}
    \caption{Local explanation for classifying a test image of a Tree Swallow. Interestingly, the 6th prototype could be detected in the test image because of the white-colored chest or because of the similarity with the curved branch. An explanation as presented by~\cite{nauta2020looks} to indicate whether color hue or shape is important, could clarify this.}
    \label{fig:local_cub_2}
\end{figure*}

\begin{figure*}[!ht]
    \centering
    \includegraphics[width=0.85\textwidth]{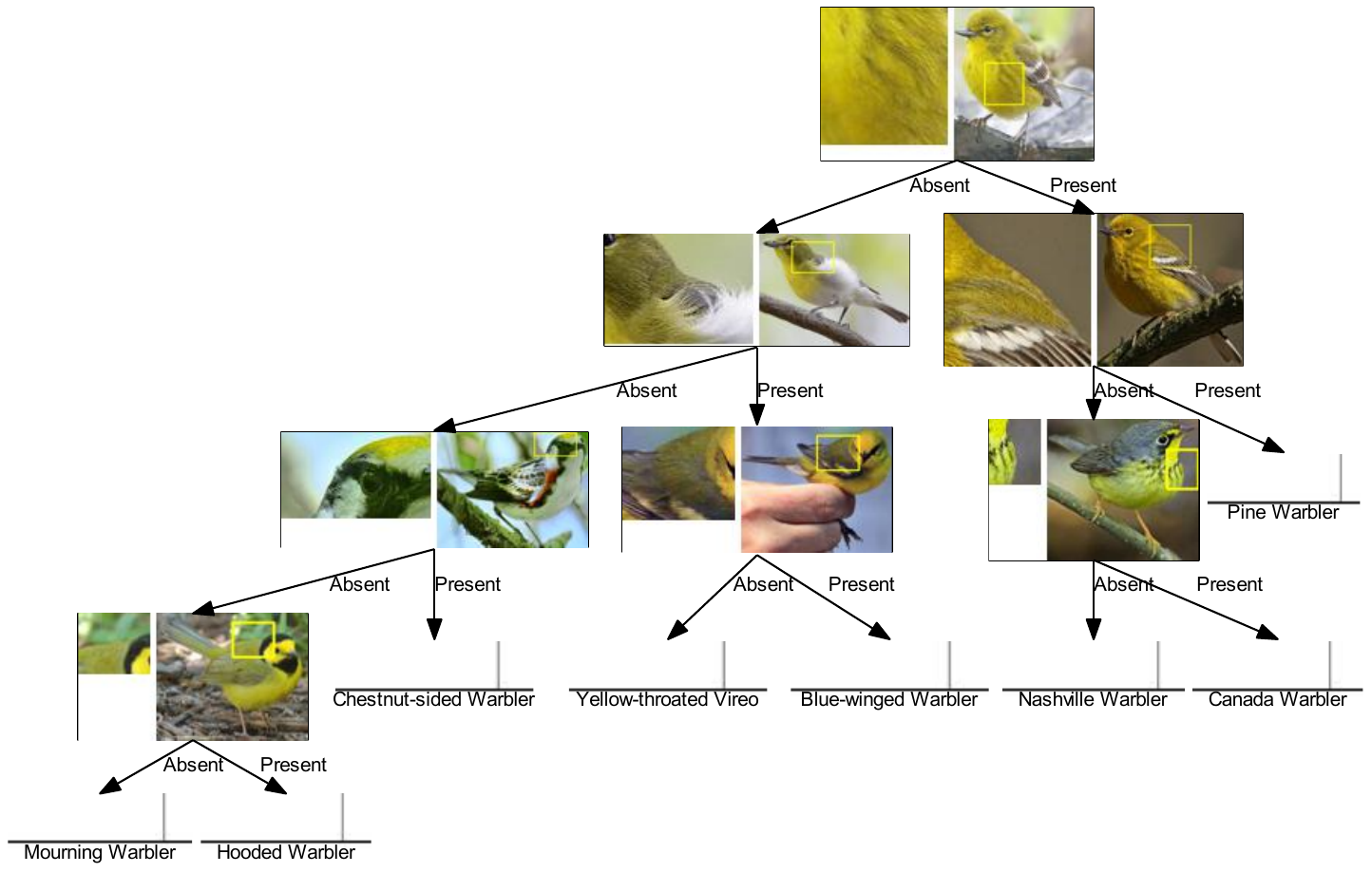}
    \caption{Subtree of an automatically visualized ProtoTree (CUB, $h=8$).  A ProtoTree hierarchically clusters similar classes, in this case Warblers.}
    \label{fig:cub_h8_example_warblers}
\end{figure*}

\begin{figure*}
\centering
\begin{minipage}[b]{.45\textwidth}
\includegraphics[width=\linewidth]{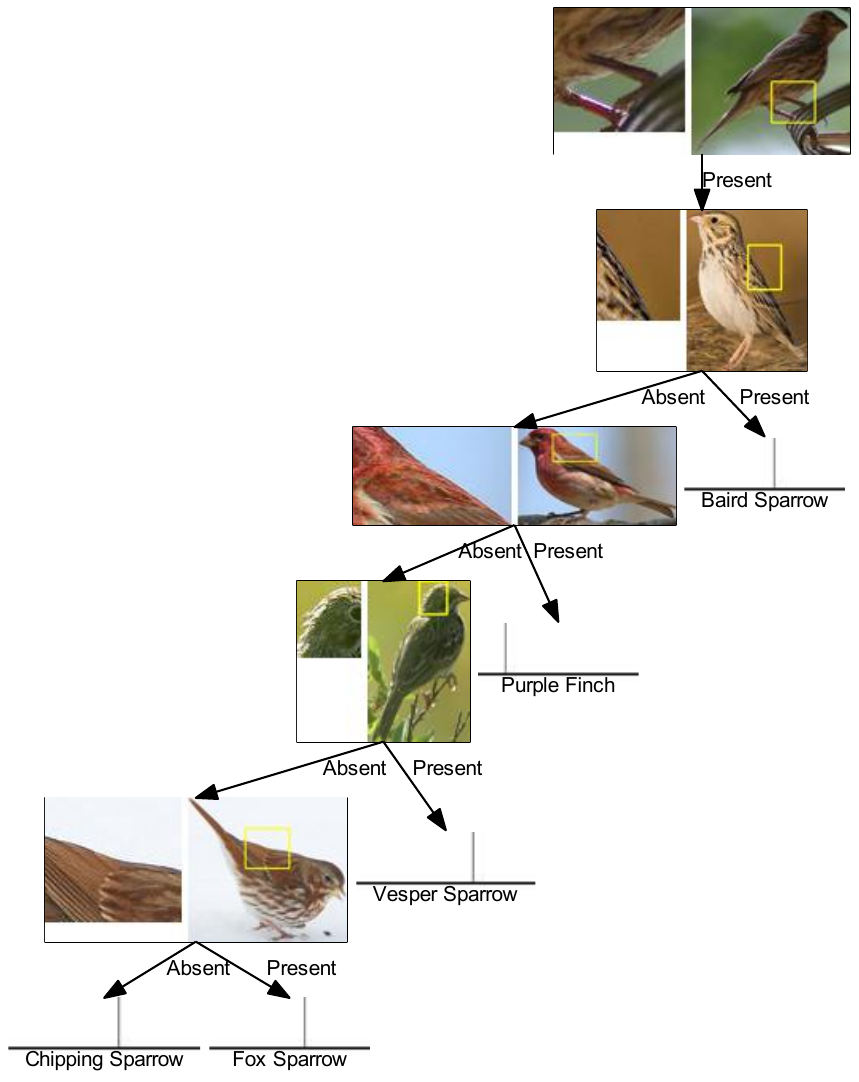}
    \caption{Subtree of a ProtoTree (CUB, $h=9$). The top node clusters birds with red legs and a light colored abdomen. Pruning can result in a deep, imbalanced tree. }
    \label{fig:cub_h8_redlegs}
\end{minipage}\qquad
\begin{minipage}[b]{.49\textwidth}
\includegraphics[width=\linewidth]{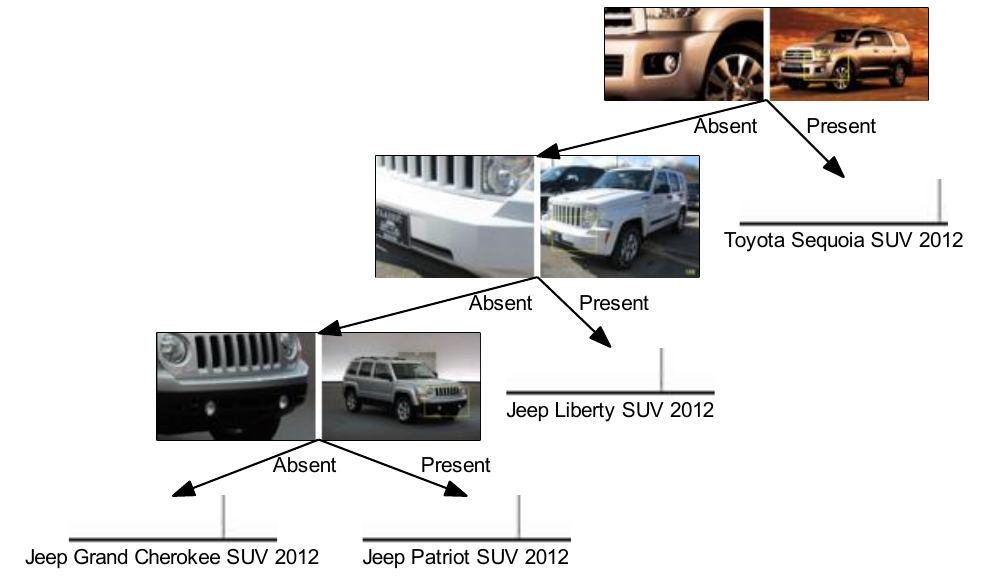}
    \caption{Subtree of a ProtoTree (CARS, $h=10$) which clusters similar SUV's. Here, pruning results in an imbalanced tree.}
    \label{fig:cars_h10_1218}
\end{minipage}
\end{figure*}

\begin{figure*}
    \centering
    \includegraphics[width=0.94\textwidth]{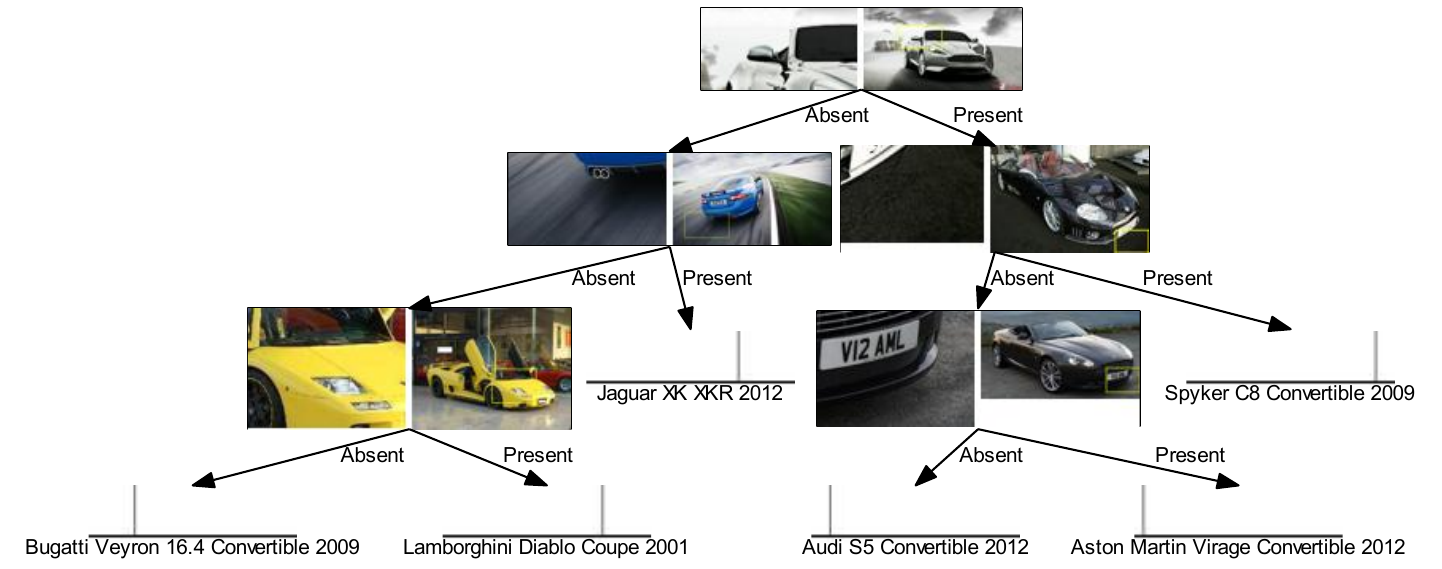}
    \caption{Subtree of a ProtoTree (CARS, $h=9$) with convertibles clustered on the right. Each internal node contains a prototype (left) and the training image from which it is extracted (right).}
    \label{fig:cars_h9_834}
\end{figure*}

\begin{figure*}
    \centering
    \includegraphics[width=\textwidth]{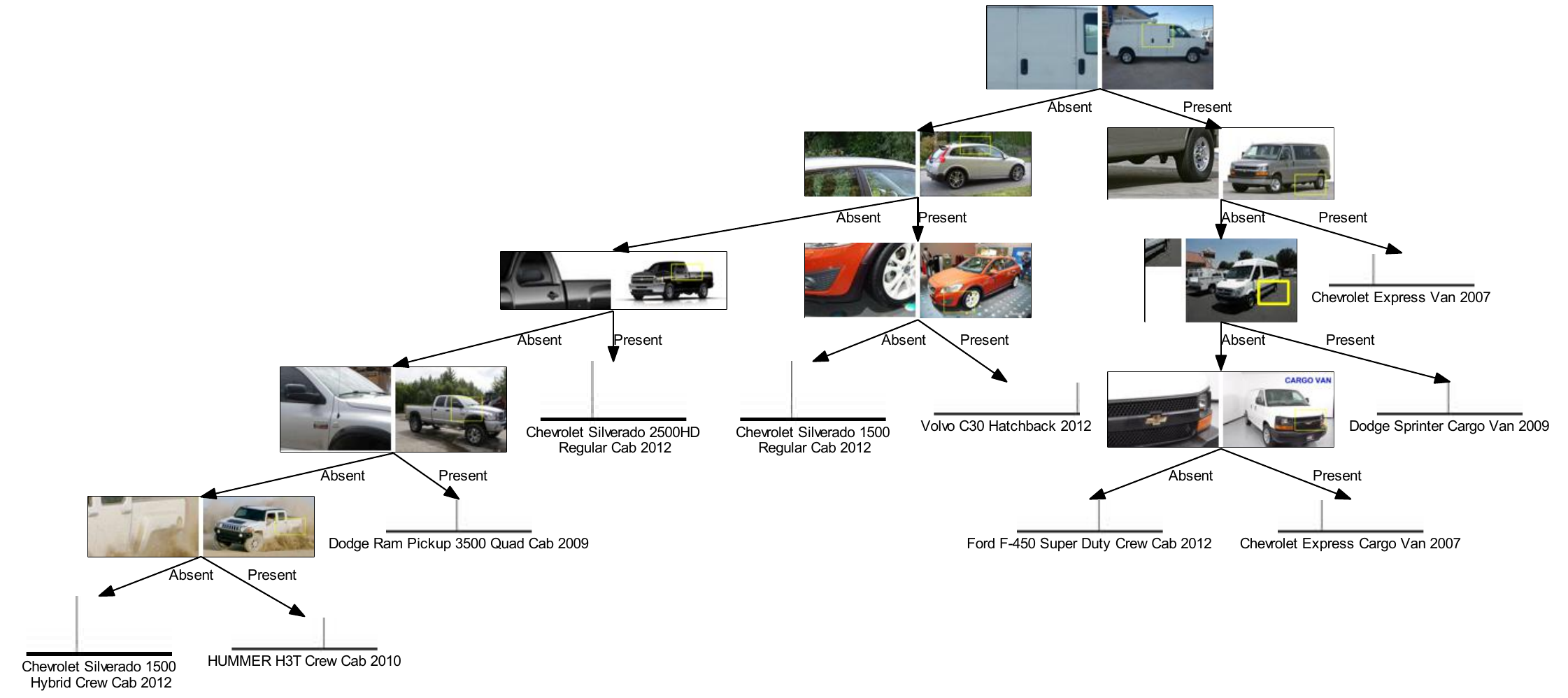}
    \caption{Subtree of a ProtoTree (CARS, $h=11$). Vans are clustered on the right, and pickup trucks on the left.}
    \label{fig:cars_h11_6}
\end{figure*}

\begin{figure*}
    \centering
    \includegraphics[width=\textwidth]{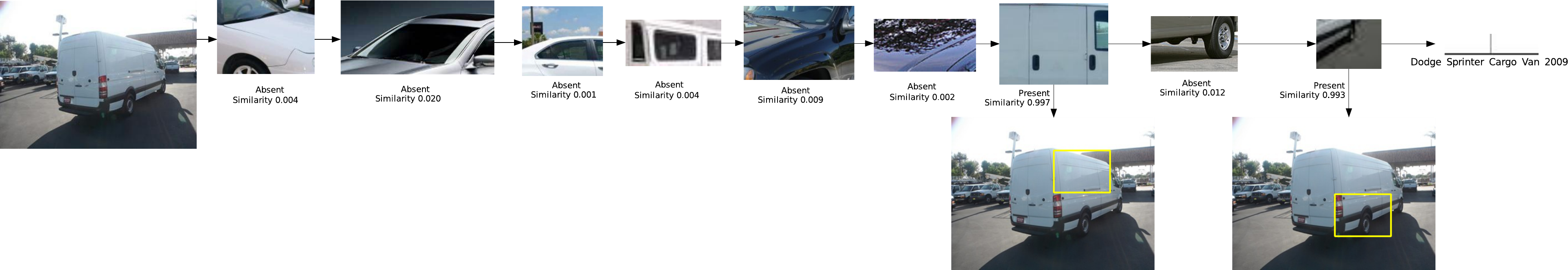}
    \caption{Local explanation for classifying a test image of a Dodge Sprinter Cargo Van 2009 (CARS, $h=11$), recognizable by the black stripe on the side.}
    \label{fig:local_dodge_cars_h11_6}
\end{figure*}

\begin{figure*}
    \centering
    \includegraphics[width=0.97\textwidth]{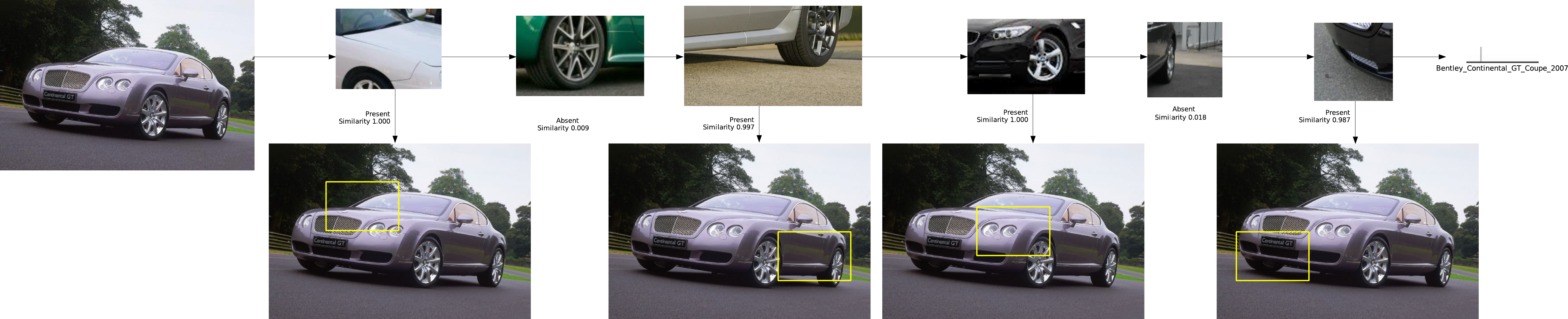}
    \caption{Local explanation for classifying a test image of a Bentley Continental GT Coupe 2007 (CARS, $h=11$).}
    \label{fig:local_bentley_cars_h11_6}
\end{figure*}

\begin{figure*}
    \centering
    \includegraphics[width=0.95\textwidth]{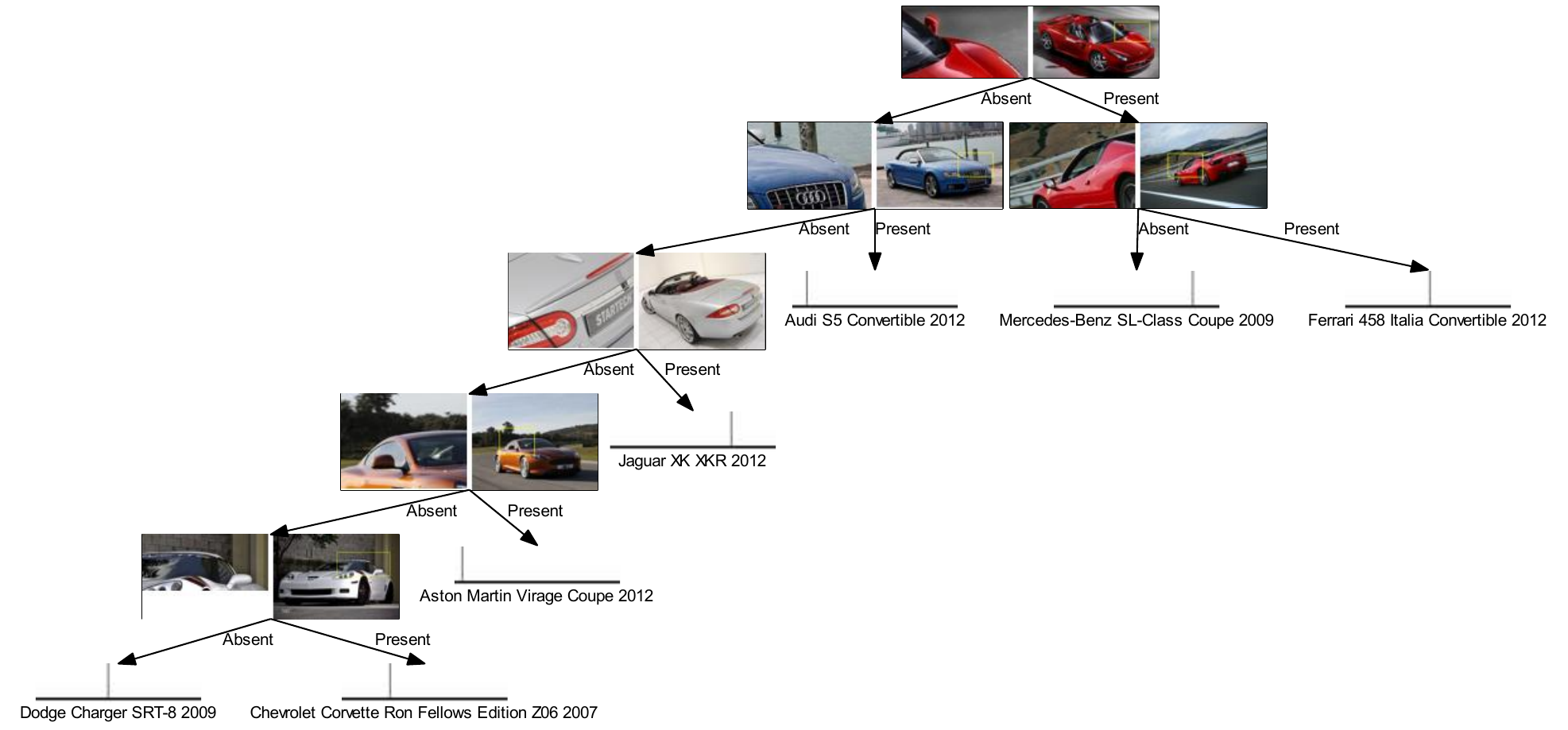}
    \caption{Subtree of a ProtoTree (CARS, $h=10$). The top node clusters two cars that are styled with similar feature lines on the hood. The Audi is recognized by its logo. }
    \label{fig:cars_h10_1792}
\end{figure*}

\begin{figure*}
    \centering
    \includegraphics[width=\textwidth]{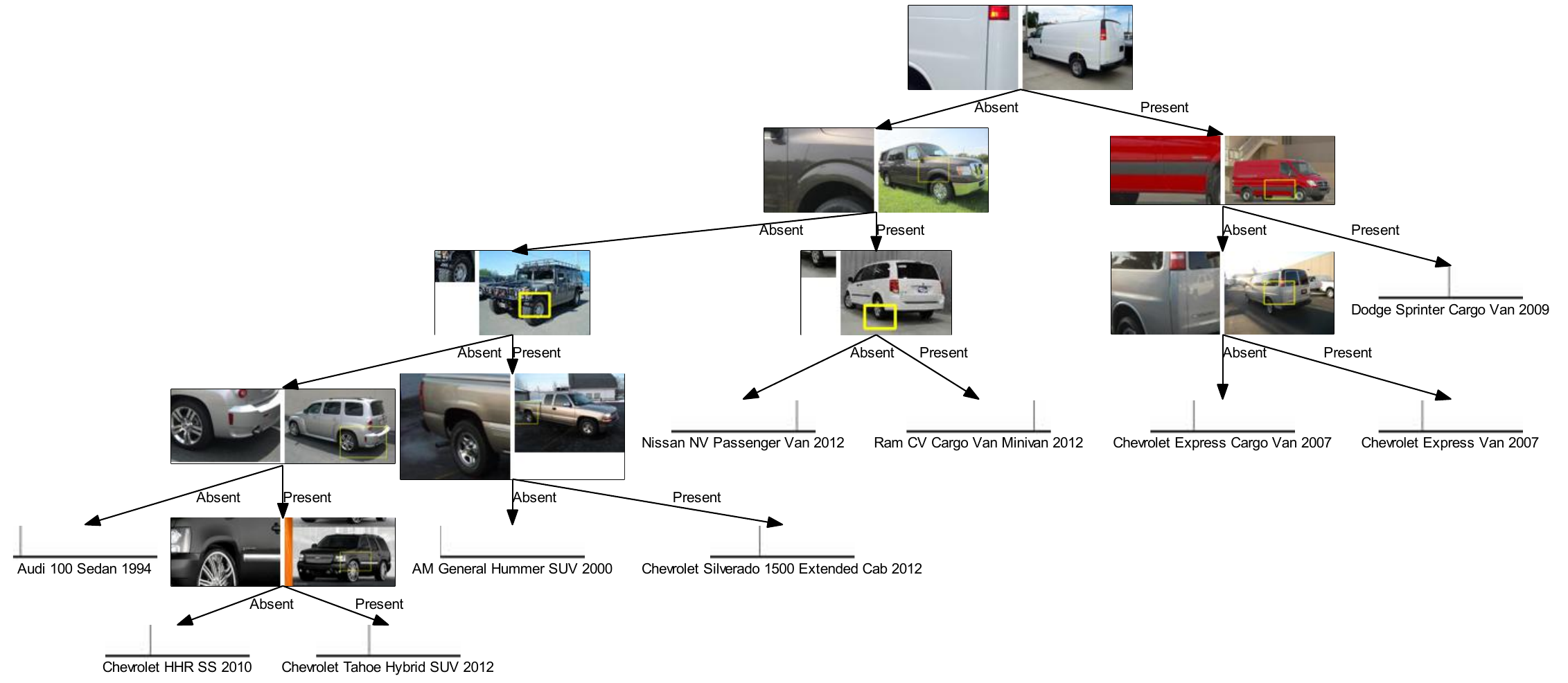}
    \caption{Subtree of a ProtoTree (CARS, $h=10$). Similar vans are clustered on the right. Chrevrolets are recognized by their distinctive back.}
    \label{fig:cars_h10_1537}
\end{figure*}

\clearpage
\twocolumn
{\small
\bibliographystyle{ieee_fullname}
\bibliography{supp_bib}
}